\documentclass[citeauthoryear]{llncs}
\usepackage{llncsdoc}
\usepackage{graphicx}
\usepackage{amsmath}
\usepackage{epstopdf}

\begin{document}
\title{Euclidean Distances, soft and spectral  Clustering \\ on Weighted Graphs}

\author{Fran\c{c}ois Bavaud}

\institute{University of Lausanne,  Department of Geography\\ Department of Computer Science and Mathematical Methods \\ÊB\^atiment Anthropole, CH-1015 Lausanne, Switzerland}

%\medskip\noindent

\maketitle

 \begin{abstract}
We define a class of Euclidean distances on weighted graphs, enabling to perform thermodynamic  soft graph clustering.  The class can be constructed form the ``raw coordinates" encountered in spectral clustering, and can be extended by means of higher-dimensional  embeddings (Schoenberg transformations). Geographical flow data, properly conditioned, illustrate the procedure as well as visualization aspects.
\end{abstract}
 

\section{Introduction}
In a nutshell (see e.g. Shi and  Malik (2000); Ng,  Jordan and Weiss (2002);  von Luxburg (2007) for a review), spectral graph clustering consists in 
\begin{enumerate}
  \item[A)] constructing a features-based similarity or affinity matrix between $n$ objects
  \item[B)]  performing the spectral decomposition of the 
normalized affinity matrix,  and representing the objects by the corresponding eigenvectors or {\em raw coordinates} 
  \item[C)]  applying a clustering algorithm on the raw coordinates.
\end{enumerate}
 
The present contribution focuses on (C)  thermodynamic clustering (Rose et al. 1990; Bavaud 2009), an aggregation-invariant  soft $K$-means clustering  based upon {\em Euclidean distances between objects}. The latter constitute {\em distances on weighted graphs}, and are constructed from the raw coordinates (B), whose form happens to be justified from presumably new considerations on equivalence between vertices  (Section \ref{tncsed}). Geographical {\em flow data} illustrate the theory (Section \ref{quatre}). Once properly symmetrized, endowed with a sensible diagonal and normalized, flows define an {\em exchange matrix} (Section \ref{deux}), that is an 
affinity matrix (A) which might be positive definite or not. 

A particular emphasis is devoted to the definition of Euclidean distances on weighted graphs and their properties (Section \ref{trois}). For instance, diffusive and chi-square distances are {\em focused}, that is zero between {\em equivalent} vertices. Commute-time and absorption distances are not focused, but their values between equivalent vertices possess an universal character. All these distances, whose relationships to the shortest-path distance on weighted graphs is partly elucidated, differ in the way eigenvalues are used to scale the raw coordinates. Allowing further {\em Schoenberg transformations} (Definition \ref{edw})
of the distances still extends the class of admissible distances on graphs, by means of a high-dimensional embedding familiar in the Machine Learning community.

\section{Preliminaries and notations}
\label{deux}
Consider $n$ objects, together with an {\em exchange} matrix $E=(e_{ij})$, that is a   $n\times n$
non-negative, symmetric  matrix, whose components add up to unity (Berger and Snell 1957). $E$ can be obtained by normalizing an affinity of similarity matrix, and defines the normalized adjacency matrix of a weighted undirected graph (containing loops in general), where $e_{ij}$ is the weight of edge $(ij)$ and $f_i=\sum_{j=1}^n e_{ij}$ is the {\em relative degree} or {\em weight} of vertex $i$, assumed strictly positive. 

\subsection{Eigenstructure}
$P=(p_{ij})$ with $p_{ij}=e_{ij}/f_i$ is the transition matrix of a reversible Markov chain, with 
stationary distribution $f$.  The $t$-step exchange matrix is $E^{(t)}=\Pi P^t$, where $\Pi$ is the diagonal  matrix containing the weights $f$. In particular, assuming the chain to be regular (see e.g. Kijima 1997)
\begin{displaymath}
E^{(0)}=\Pi\qquad\qquad
E^{(2)}=E\Pi^{-1}E\enspace .
\qquad\qquad
E^{(\infty)}=ff'
\end{displaymath}
$P$ is similar to the symmetric, {\em normalized exchange matrix} $\Pi^{-\frac12}E\Pi^{-\frac12}$ (see e.g. Chung 1997), and share the same eigenvalues $1=\lambda_0\ge\lambda_1\ge \lambda_2\ge\ldots
\lambda_{n-1}\ge-1$. It is well-known that the second eigenvalue $\lambda_1$ attains its maximum value 1 iff the graph contains disconnected components, and $\lambda_{n-1}=-1$ iff the graph is bipartite. We note $U'\Lambda U$ the spectral decomposition of the normalized exchange matrix, where $\Lambda$ is diagonal and contains the eigenvalues, and $U=(u_{i\alpha})$ is orthonormal and contains the normalized eigenvectors. In particular, $u_0=\sqrt{f}$ is the eigenvector corresponding to the trivial eigenvalue $\lambda_0=1$. Also, the spectral decomposition of higher-order  exchange matrices reads 
$\Pi^{-\frac12}E^{(t)}\Pi^{-\frac12}=U\Lambda^tU'$.

\subsection{Hard and soft partitioning}
\label{hsop}
A {\em soft partition} of the $n$ objects into $m$ groups is specified by a $n\times m$ {\em membership matrix} $Z=(z_{ig})$, whose components (obeying $z_{ig}\ge0$ and $\sum_{g=1}^m z_{ig}=1$) quantify the membership degree of object $i$ in group $g$. The relative  {\em volume} of  group $g$  is $\rho_g=\sum_i f_i z_{ig}$.  The  components $\theta_{gh}=\sum_i f_i z_{ig}z_{ih}$ of the 
$m\times m$ matrix $\Theta=Z'\Pi Z$   measure the {\em overlap} between groups $g$ and $h$. In particular, $\theta_{gg}/\rho_g\le~1$  measures the {\em hardness} of group $g$. 
The  components $a_{gh}=\sum_{ij} e_{ij} z_{ig}z_{jh}$ of the 
$m\times m$ matrix $A=Z' E Z$   measure the {\em association} between groups $g$ and $h$.

% Inequality 
% $\sum_{ij}e_{ij}(z_{ig}-z_{jh})^2\ge0 $ demonstrates $2 a_{gh}\le \theta_{gg}+\theta_{hh}$. 

A group $g$ can also be specified by the objects it contains, namely by the {\em distribution} 
$\pi^g$ with components $\pi_i^g=f_iz_{ig}/\rho_g$, obeying $\sum_i \pi_i^g=1$ by construction.
The object-group {\em mutual information} 
\begin{eqnarray*}
%\hspace{-0.3cm}
%\mbox{\small $
\mbox{\small $I(O,Z) = H(O)+H(Z)-H(O,Z) 
 = -\sum_if_i\ln f_i-\sum_g\rho_g\ln\rho_g+
\sum_{ig}f_iz_{ig}\ln(f_iz_{ig})$}
\end{eqnarray*}
measures the  object-group dependence or cohesiveness  (Cover and Thomas 1991).

A partition is {\em hard}  if each object belongs to an unique group, that is if the memberships are of the form $z_{ig}=I(i\in g)$, or equivalently if  $z_{ig}^2=z_{ig}$ for all $i,g$, or equivalently if $\theta_{gg}=\rho_g$ for all $g$,  or still equivalently if the {\em overall softness} $H(Z|O)=H(Z)-I(O,Z)$ takes on its minimum value of zero.

Also, $H(O)\le \ln n$, with equality iff $f_i=1/n$, that is if the graph is regular. 
 
\subsection{Spectral versus soft  membership  relaxation}
In their presentation of the Ncut-driven spectral clustering,  Yu and Shi (2003) (see also Nock et al. 2009) determine the hard  $n\times m$ membership $Z$ maximizing 
\begin{displaymath}
\epsilon[Z]=\sum_{g=1}^m \frac{a_{gg}}{\rho_g} = \sum_g \frac{a_{gg}}{\theta_{gg}}=\mbox{tr}(X' E X)\qquad \qquad \mbox{where}\quad X[Z]=Z\: \Theta^{-\frac12}[Z]
\end{displaymath}
under the constraint $X'\Pi X=I$. Relaxing the hardness  and non-negativity  conditions, they show the 
solution  to be $\epsilon[Z_0]=1+\sum_{\alpha=1}^{m-1}\lambda_\alpha$, attained with an 
optimal ``membership"   of the form $Z_0=X_0R\Theta^{\frac12}$ where $R$ is any orthonormal $m\times m$ matrix and $X_0=({\bf 1},x_1,\ldots,x_\alpha,\ldots,x_{m-1})$ is the $n\times m$ matrix formed by the unit vector followed by  of the  first {\em raw coordinates} (Sec.  \ref{tncsed}). The above spectral  relaxation of the memberships, involving the eigenstructure of the normalized exchange matrix, completely differs from the soft membership relaxation which will be used in Section \ref{thermo}, preserving positivity and normalization of $Z$.
 
\section{Euclidean distances on weighted graphs}
\label{trois}

\subsection{Squared Euclidean distances}
Consider a collection of $n$ objects together with an associated  pairwise distance. A successful clustering consists in partitioning the objects into $m$ groups, such that the average distances between objects belonging to the same (different) group are small (large). The most 
tractable pairwise distance is, by all means, the {\em squared Euclidean distance} $D_{ij}=\sum_{c=1}^q (x_{ic}-x_{jc})^2$, where $x_{ic}$ is the coordinate of object $i$ in dimension $c$. Its virtues follow from {\em Huygens principles}
\begin{equation}
\label{huy}
\sum_{j} p_j D_{ij}=D_{ip}+\Delta_p\qquad\qquad
\Delta_p=\sum_j p_j D_{jp}=\frac12\sum_{ij}p_i p_j D_{ij}
\end{equation}
where $p_i$ represents a (possibly non positive) {\em signed distribution}, i.e. obeying $\sum_ip_i=1$, $D_{ip}$ is the squared Euclidean distance between $i$ and the centroid of coordinates 
$\bar{x}_{pc}=\sum_i p_i x_{ic}$, and $\Delta_p$ the average pairwise distance or {\em inertia}. Equations (\ref{huy}) are easily checked using the coordinates, although the  latter do {\em  not}  explicitly appear in the formulas. To that extent, squared Euclidean distances enable a  feature-free formalism, a property shared with the kernels of Machine Learning, and to the ``kernel trick" of Machine Learning amounts an equivalent  ``distance trick" (Sch{\"o}lkopf  2000; Williams 2002), as expressed by the 
well-known  {\em Classical Multidimensional Scaling} (MDS) procedure. Theorem \ref{theomds} below presents a weighted version (Bavaud 2006), generalizing the  uniform MDS  procedure (see e.g. Mardia et al. 1979). Historically, MDS has been developed from the independent contributions of 
Schoenberg (1938b) and Young and Householder (1938). The algorithm has been  popularized  by Torgeson (1958) in Data Analysis.

\begin{theorem}[weighted classical MDS]
\label{theomds}
The dissimilarity square  matrix $D$ between $n$ objects with weights $p$ is a squared Euclidean distance iff the {\em scalar product matrix} 
$B=-\frac12 HDH'$ is (weakly) positive definite (p.d.), where $H$ is the $n \times n$ centering matrix
with components $h_{ij}=\delta_{ij}-p_j$. By construction, $B_{ij}=-\frac12(D_{ij}-D_{ip}-D_{jp})$ and $D_{ij}=B_{ii}+B_{jj}
-2B_{ij}$.
The object  coordinates  can be reconstructed as
$x_{i\beta}=\mu_\beta^\frac12 p_i^{-\frac12}v_{i\beta}$ for $\beta=1,2,\ldots$, where the
$\mu_\beta$ are the decreasing eigenvalues and the  $v_{i\beta}$  are the eigenvectors occurring in the spectral decomposition $K=VMV'$ of the {\em weighted scalar product} or {\em kernel} $K$ with components
$K_{ij}=\sqrt{p_ip_j}B_{ij}$. This reconstruction provides the optimal low-dimensional reconstruction of the inertia associated  to $p$
\begin{displaymath}
\Delta=\frac12 \sum_{ij}p_i p_j D_{ij}=\mbox{tr}(K)=\sum_{\beta\ge1}\mu_\beta\enspace .
\end{displaymath}
Also, the Euclidean (or not) character of $D$ is independent of the choice of $p$. 
\end{theorem}

\subsection{Thermodynamic clustering}
\label{thermo}
Consider the overall objects weight $f$, defining a  centroid denoted by $0$, together with $m$ soft groups defined by their distributions $\pi^g$ for $g=1,\ldots,m$, with associated centroids denoted by $g$. By (\ref{huy}), the overall inertia decomposes as 
\begin{eqnarray*}
\mbox{\small $\Delta=\sum_i f_i D_{i0}=\sum_{ig}f_i z_{ig}D_{i0}=\sum_g \rho_g  \sum_i \pi^g_i D_{i0}=
\sum_g \rho_g[D_{g0}+\Delta_g]=\Delta_B+\Delta_W$}
\end{eqnarray*}
where $\Delta_B[Z]=\sum_g \rho_g D_{g0}$ is the between-groups inertia, and 
$\Delta_W[Z]=\sum_g \rho_g \Delta_g$ the within-groups inertia. The optimal clustering is then provided by the $n\times m$ membership matrix $Z$ minimizing $\Delta_W[Z]$, or  equivalently  maximizing 
$\Delta_B[Z]$. The former functional can be shown  to be {\em concave} in $Z$ (Bavaud 2009), implying the minimum to be attained for {\em hard} clusterings. 

Hard clustering is notoriously computationally  intractable and some kind of regularization is required. Many authors (see e.g. Huang and Ng (1999) or Filippone et al. (2008)) advocate  the use of the {\em $c$-means clustering}, involving a power transform of the memberships. Despite its efficiency and popularity, the $c$-means algorithm actually 
suffers from a serious formal defect, questioning its very logical foundations: its objective function is indeed {\em not aggregation-invariant}, that is generally changes when two groups $g$ and $h$ supposed  equivalent  in the sense $\pi^g=\pi^h$ are merged into a single group $[g\cup h]$ with membership $z_{i[h\cup g]}=z_{ih}+z_{jh}$ (Bavaud 2009). 

An alternative, aggregation-invariant regularization is provided by the {\em thermodynamic clustering}, minimizing over  $Z$ the {\em free energy} $F[Z]=\Delta_W[Z]+T I[Z]$, where 
$I[Z]\equiv I(O,Z)$ is the  objects-groups mutual information and $T>0$ the {\em temperature} (Rose et al. 1990; Rose 1998; Bavaud 2009). The resulting membership is determined iteratively through 
\begin{equation}
z_{ig}=\frac{\rho_g\: \exp(-D_{ig}/T)}{\sum_{h=1}^m\rho_h\: \exp(-D_{ih}/T)}
\label{tcl}
\end{equation}
and converges towards a local minimum of the free energy. Equation (\ref{tcl}) amounts to fitting Gaussian clusters in the framework of {\em model-based clustering}. 
%Letting progressively $T\to0$ in a subsequent phase (``simulated annealing")  yields a hard local  
% minimum of $\Delta_W[Z]$. 

\subsection{Three nested classes of squared Euclidean distances}
\label{tncsed}
Equation  (\ref{tcl}) solves the K-way soft graph clustering problem, given of course the availability of  a sound  class of squared Euclidean distances on weighted graphs. Definitions \ref{ndw} and \ref{edw} below
%,  based upon spectral considerations, 
seem to solve the latter issue. 

Consider a graph possessing two distinct but equivalent vertices in the sense their relative exchange is identical  with the other vertices (including themselves). Those vertices  somehow stand as duplicates of the same object, and one could as a first attempt require their distance to be zero. 

\begin{definition}[Equivalent vertices; focused distances]
\label{fod}
Two distinct vertices $i$ and $j$ are {\rm equivalent}, noted $i\sim j$, if $e_{ik}/f_i = e_{jk}/f_j$ for all $k$.
A distance is {\rm focused}  if $D_{ij}=0$ for $i\sim j$. 
\end{definition}

\begin{proposition}
\label{equ}
$i\sim j$ iff $x_{i\alpha}=x_{j\alpha}$ for all $\alpha\ge1$ such that  $\lambda_\alpha\neq 0$, where  $x_{i\alpha}=u_{i\alpha}/\sqrt{f_i}$ is the {\rm raw coordinate} of vertex $i$ in dimension $\alpha$.  \end{proposition} 

The proof directly follows from the substitution $e_{ik}\to f_i e_{jk}/f_j$ in the  identity 
$\sum_k f_i^{-\frac12}e_{ik}f_k^{-\frac12} u_{k\alpha}=\lambda_\alpha u_{i\alpha}$. Note that
the condition trivially holds for the trivial eigenvector $\alpha=0$, in view of $f_i^{-\frac12}u_{i0} \equiv 1$ for all $i$. It also holds trivially for the ``completely connected" weighted graph $e_{ij}^{(\infty)}=f_if_j$, where all vertices are equivalent, and all eigenvalues are zero, except the trivial one.

Hence, any expression of the form
%, say,  $d_{ij}=\sum_{\alpha\ge1}g_\alpha|f_i^{-\frac12}u_{i\alpha}-f_j^{-\frac12}u_{j\alpha}|$ with $g_%\alpha\ge0$ constitutes a $L_1$ distance (see e.g. ***cutbook*** et *** Fichet et crichley?**** ) 
% vanishing between equivalent vertices, and any expression of the form 
$D_{ij}=\sum_{\alpha\ge1}g_\alpha(f_i^{-\frac12}u_{i\alpha}-f_j^{-\frac12}u_{j\alpha})^2$ with $g_\alpha\ge0$ constitutes an admissible  squared Euclidean distance, obeying $D_{ij}=0$ for $i\sim j$, {\em provided $g_\alpha=0$ if $\lambda_\alpha=0$}. The quantities $g_\alpha$ are non-negative, but otherwise arbitrary; however, it is natural to require the latter to depend upon the sole parameters at disposal, namely the eigenvalues, that is to set $g_\alpha=g(\lambda_\alpha)$.

\begin{definition}[Focused and Natural Distances on Weighted Graphs]
\label{ndw}
Let $E$ be the exchange matrix associated to a weighted graph, and define
$E^s:=\Pi^{-\frac12}(E-E^{(\infty)})\Pi^{-\frac12}$,  the {\em standardized} exchange matrix. 
 The class of {\rm  focused squared Euclidean distances on weighted graphs}Ê is
\begin{eqnarray*}
D_{ij} = B_{ii}+B_{jj}-2B_{ij},\quad  \mbox{where}\quad  B   =   \Pi^{-\frac12}K\Pi^{-\frac12} \quad\mbox{and}\quad
K=g(E^s)
\end{eqnarray*}
where $g(\lambda)$ is any non-negative sufficiently regular real function with $g(0)=0$. Dropping the requirement $g(0)=0$ defines the more general  class of {\rm natural} squared Euclidean distances on weighted graphs.

If   $g(1)$ is finite, $K$ can also  be defined as $K=g(\Pi^{-\frac12}E\Pi^{-\frac12})=Ug(\Lambda)U'$.
\end{definition}

%A few comments may help grasping Definition \ref{ndw}: 
First, note the standardized exchange matrix to result from a ``centering" (eliminating the trivial eigendimension)  followed by a ``normalization":
\begin{equation}
\label{sta}
e_{ij}^s=\frac{e_{ij}-f_if_j}{\sqrt{f_if_j}}=\sum_{\alpha\ge1}\lambda_\alpha u_{i\alpha}u_{j\alpha}
\enspace .
\end{equation} 
Secondly,  $B$ is the  matrix of scalar products appearing in Theorem \ref{theomds}. The resulting  optimal reconstruction coordinates are $\sqrt{g(\lambda_\alpha)}\: x_{i\alpha}$, where the quantities  $x_{i\alpha}=f_i^{-\frac12}u_{i\alpha}$ are the {\em raw coordinates} of vertex $i$ in dimension $\alpha=1,2,\ldots$ appearing in Proposition \ref{equ} - which yields a 
%rather strong and 
general rationale for their widespread use in clustering and  low-dimensional visualization.
% (*** citer plein de monde***). 
 Thirdly, the matrix $g(E^s)$ can be defined, for $g(\lambda)$ regular enough, as the power expansion in 
$(E^s)^t$ with  coefficients given by the power expansion of $g(\lambda)$ in $\lambda^t$, for 
$t=0,1,2,\ldots$. Finally, the two variants of $B$ appearing  in Definition \ref{ndw} are identical up to a matrix $g(1) {\bf 1}_n{\bf 1}_n'$, leaving $D$ unchanged.  

If $g(1)=\infty$, the distance between vertices belonging to distinct irreducible components becomes infinite: recall the graph to be disconnected iff $\lambda_1=1$. Such distances will be referred to as {\em irreducible}.

Natural distances are in general not focused. The distances between equivalent vertices are however  {\em universal}, that is independent of the details of the graph or of the associated distance (Proposition \ref{unf}). To demonstrate this property, consider first an equivalence class $J:=\{k\: |\: k\sim j\}$ containing at least two equivalent vertices. 
Aggregating the vertices in $J$  results in a new $\tilde{n}\times \tilde{n}$ 
 exchange matrix $\tilde{E}$ with $\tilde{n}=(n-|J|-1)$, with components $\tilde{e}_{JJ}=\sum_{ij\in J}e_{ij}$, $\tilde{e}_{Jk}=\tilde{e}_{kJ}=\sum_{j\in J}e_{jk}$ for $k\notin J$ and  $\tilde{f}_J=\sum_{j\in J}f_j$, 
the other components remaining unchanged.

\begin{proposition}
\label{unf}
Let $D$ be a natural distance and consider a graph possessing an equivalence class $J$ of size $|J|\ge 2$. Consider  two distinct elements $i\sim j$ of $J$  and let $k\notin J$. Then 
\begin{displaymath}
D_{ij}=g(0)\: (\frac{1}{f_i}+\frac{1}{f_j})\qquad\qquad 
D_{jJ}=g(0)\:(\frac{1}{f_i}-\frac{1}{\tilde{f}_J})
\qquad\qquad
\Delta_J=g(0)\:\frac{|J|-1}{\tilde{f}_J}\enspace .
\end{displaymath}
Moreover, the Pythagorean relation $D_{kj}=D_{kJ}+D_{jJ}$ holds.
 \end{proposition} 

\vspace{-0.2cm}\noindent {\bf Proof:} consider the eigenvalues $\tilde{\lambda}_\beta$ and eigenvectors $\tilde{u}_\beta$, associated to the aggregated graph $\tilde{E}$, for $\beta=0,\ldots, \tilde{n}$. One can check that, 
 due to the collinearity generated by the $|J|$ equivalent vertices,
 \begin{enumerate}
  \item[$\bullet$]  $\tilde{n}$ among the original eigenvalues $\lambda_\alpha$ coincide with the set of aggregated eigenvalues 
 $\tilde{\lambda}_\beta$ (non null in general),  with corresponding eigenvectors  $u_{j\beta}=
 f_j^\frac12\tilde{f}_J^{-\frac12}\tilde{u}_{J\beta}$ for $j\in J$ and $u_{k\beta}=
\tilde{u}_{k\beta}$ for $k\notin J$
\item[$\bullet$]  $|J-1|$ among the original eigenvalues $\lambda_\alpha$ are zero. Their corresponding eigenvectors are of the form $u_{j\gamma}=h_{j\gamma}$ for $j\in J$ and 
$u_{k\gamma}=0$ for $k\notin J$, where the $h_\gamma$ constitute the $|J|-1$ columns  of an orthogonal $|J|\times |J|$ matrix, the remaining column being
$( f_j^\frac12\tilde{f}^{-^\frac12}_J)_{j\in J}$.
\end{enumerate}
Identities in Proposition \ref{unf} follow by substitution. For instance, 
\begin{displaymath}
D_{ij}=\sum_{\beta=1}^{\tilde{n}}g(\lambda_\beta)(\frac{u_{i\beta}}{\sqrt{f_i}}-\frac{u_{j\beta}}{\sqrt{f_j}})^2
+g(0)\sum_{\gamma=1}^{|J|-1}(\frac{h_{i\gamma}}{\sqrt{f_i}}-\frac{h_{j\gamma}}{\sqrt{f_j}})^2=g(0)\: (\frac{1}{f_i}+\frac{1}{f_j})\enspace .
\end{displaymath}

 General at it is,  the class  of  squared Euclidean distances on weighted graphs of Definition \ref{ndw} can still be extended: a wonderful result of Schoenberg (1938a), still apparently little known in the Statistical and Machine Learning community (see however the references in Kondor and Lafferty (2002); Hein et al. (2005))  
 %; Bavaud (2010)
asserts that the componentwise correspondence $\tilde{D}_{ij}=\phi(D_{ij})$ transforms any  squared Euclidean distance $D$ into another  squared Euclidean  distance $\tilde{D}$, provided that 
\begin{enumerate}
  \item[i)] $\phi(D)$ is positive with $\phi(0)=0$
  \item[ii)]  odd derivatives $\phi'(D)$, $\phi'''(D)$,... are positive
  \item[iii)]  even derivatives $\phi''(D)$, $\phi''''(D)$,... are negative.
\end{enumerate}
For example, $\phi(D)=D^a$ (for $0<a\le1$) and $\phi(D)=1-\exp(- b D)$ (for $b>0$) are instances of such {\em Schoenberg transformations} (Bavaud 2010).

\begin{definition}[Extended Distances on Weighted Graphs]
\label{edw}
The class of {\rm  extended  squared Euclidean distances on weighted graphs}Ê is
\begin{eqnarray*}
\tilde{D}_{ij}=\phi(D_{ij})
\end{eqnarray*}
where $\phi(D)$ is a {\em Schoenberg transformation} (as specified above), and $D_{ij}$ is  a natural squared Euclidean distance 
 associated to the weighted graph $E$,
in the sense of Definition \ref{ndw}. 
\end{definition} 

\subsection{Examples of distances on weighted graphs}
\subsubsection{The chi-square distance}
The choice $g(\lambda)=\lambda^2$ entails, together with (\ref{sta}) 
\begin{displaymath}
\Delta=\mbox{tr}(K)=\mbox{tr}((E^s)^2)=\sum_{ij}\frac{(e_{ij}-f_if_j)^2}{ f_if_j}=\chi^2
\end{displaymath}
which is the familiar chi-square  measure of the overall rows-columns dependency in a (square) contingency table, with 
distance  
$D^\chi_{ij}=\sum_{k}f_k^{-1}(f_i^{-1}e_{ik}-f_j^{-1}e_{jk})^2$, well-known in the {\em Correspondence Analysis} community (Lafon and Lee 2006; Greenacre 2007 and references therein). Note that $D^\chi_{ij}=0$ for $i\sim j$, as it must.

\subsubsection{The  diffusive distance}
The choice $g(\lambda)=\lambda$ is legitimate, provided the exchange matrix is purely {\em diffusive}, that is p.d. Such are typically the graphs resulting from inter-regional migrations  (Sec. \ref{quatre}) or social mobility tables (Bavaud  2008). As most people do not change place or status during the observation time, 
 the exchange matrix is strongly  dominated by its diagonal, and hence p.d.
 
Positive definiteness also occurs for graphs defined from the  affinity matrix $\exp(-\beta D_{ij})$ (Gaussian kernel), as in  Belkin and  Niyogi (2003), among many others. Indeed, distances derived from the Gaussian kernel provide a prototypical example of Schoenberg transformation  (see Definition \ref{edw}). By contrast, the affinity $I(D_{ij}\le~\epsilon^2)$ used by Tenenbaum et al. (2000) is not p.d.

The corresponding distance, together with the inertia, plainly read  
\begin{displaymath}
D^{\mbox{\small\it dif}}_{ij}=\frac{e_{ii}}{f_i^2}+\frac{e_{jj}}{f_j^2}
-2\frac{e_{ij}}{f_if_j}\qquad\qquad \Delta^{\mbox{\small\it dif}}=\sum_i\frac{e_{ii}}{f_i}-1\enspace .
\end{displaymath}
%The latter constitutes an (im)mobility index  (Shorrocks 1978; Bavaud  2008). 

\subsubsection{The ``frozen" distance}
The choice $g(\lambda)\equiv 1$ produces, for any graph, a result identical to the application 
of  any function $g(\lambda)$ (with $g(1)=1$) to 
the  purely diagonal ``frozen" graph $E^{(0)}=\Pi$, 
namely (compare with  Proposition \ref{unf}):
\begin{displaymath}
D_{ij}^{\mbox{\small\it fro}}=\frac{1}{f_i}+\frac{1}{f_j}\quad\mbox{for $i\neq j$}
\qquad \qquad 
D_{i0}^{\mbox{\small\it fro}}=\frac{1}{f_i}-1
\qquad \qquad
\Delta^{\mbox{\small\it fro}}=n-1
\enspace .
\end{displaymath}
%Completely disconnected graphs are arguably of limited interest in clustering; in taxonomy, the frozen 
%distance might however help in defining  {\em weighted Jaccard dissimilarites} between individuals 
%(see e.g. Dunn and Everitt 1982), increasing with the rarity of the characteristics possessed by the 
% individuals. 
 This ``star-like" distance (Critchley and Fichet 1994) is embeddable in a tree.

\subsubsection{The average commute time distance}
The  choice $g(\lambda)=(1-\lambda)^{-1}$ corresponds to the {\em average commute time distance};  
see Fouss et al. (2007) for a review and recent results. The amazing fact that the latter constitutes a squared Euclidean distance has only be recently explicitly recognized as such, although the key ingredients were at disposal ever since the seventies. 

Let us sketch a derivation of this result: on one hand, consider a random walk on the graph with probability transition matrix $P=\Pi^{-1}E$, and 
let $T_j$ denotes the first time the chain hits  state $j$. The average time  to go from $i$ to $j$ is 
$m_{ij}=E_i(T_j)$, with $m_{ii}=0$, where $E_i(.)$ denotes the expectation for a random walk started in  $i$. Considering the state following  $i$ yields for $i\neq j$ the relation $m_{ij}=1+\sum_kp_{ik}m_{kj}$, with solution (Kemeny and Snell (1976);  Aldous and Fill, draft chapters)
$m_{ij}=(y_{jj}-y_{ij})/f_j$, where $Y=\Pi^{-1}\sum_{t\ge0}(E^{(t)}-E^{(\infty)})=(E^{(0)}-E+E^{(\infty)})^{-1}\Pi$ is the so-called {\em fundamental matrix} of the Markov chain. On the other hand, Definition 
\ref{ndw} yields $K=(I-E^s)^{-1}=\Pi^{\frac12}(E^{(0)}-E+E^{(\infty)})^{-1}\Pi^{\frac12}=
\Pi^{\frac12}Y\Pi^{-\frac12}$, and thus $B=Y\Pi^{-1}=\Pi^{-1}Y$. Hence
\begin{displaymath}
D^{\mbox{\small\it com}}_{ij}=B_{ii}+B_{jj}-2B_{ij}=\frac{y_{ii}}{f_i}+\frac{y_{jj}}{f_j}-\frac{y_{ij}}{f_j}-\frac{y_{ij}}{f_j}=
m_{ij}+m_{ji}
\end{displaymath}
which is the  average time to go from $i$ to $j$ and back to $i$, as announced. 

Consider, for future use, the {\em Dirichlet form} ${\cal E}(y)=\frac12\sum_{ij}e_{ij}(y_i-y_j)^2$, and denote by
$y^0$  the solution of the  ``electrical" problem $\min_{y\in C_{ij}}{\cal E}(y)$, where $C_{ij}$ denotes the set of vectors $y$ such that  $y_i=1$ and $y_j=0$. Then $y_k^0=P_k(T_i<T_j)$,  where $P_k(.)$ denotes the probability for a random walk started at  $k$. Then $D^{\mbox{\small\it com}}_{ij}=1/{\cal E}(y^0)$ (Aldous and Fill, chapter 3).

\subsubsection{The shortest-path distance}
Let  $\Gamma_{ij}$ be the set of paths with extremities $i$ and $j$, where a  path $\gamma\in \Gamma_{ij}$ consists of a succession of consecutive unrepeated edges denoted by $\alpha=(k,l)\in\gamma$, whose weights $e_\alpha$ represent {\em conductances}. Their inverses are {\em resistances}, whose sum is to be minimized by the {\em shortest path} $\gamma^0\in \Gamma_{ij}$ (not necessarily unique) on the weighted graph $E$. This setup generalizes the unweighted graphs framework, and defines the {\em shortest path distance}
\begin{displaymath}
D^{\mbox{\small\it sp}}_{ij}=\min_{\gamma\in \Gamma_{ij}}\sum_{\alpha\in \gamma}\frac{1}{e_\alpha}
\enspace .
\end{displaymath}
We believe the following result to be new - although its proof simply combines a classical result 
published in the fifties (Beurling and Deny 1958) with the above ``electrical" characterization of the average commute time distance.

\begin{proposition}
\label{beur}
$D^{\mbox{\small\it sp}}_{ij}\ge D^{\mbox{\small com}}_{ij}$ with equality for all $i,j$ iff $E$ is a weighted tree.
\end{proposition} 

\vspace{-0.2cm}\noindent{\bf Proof:} let $\gamma^0\in \Gamma_{ij}$ be the shortest-path between $i$ and $j$. Consider a vector $y$ and define $dy_\alpha=y_l-y_k$ for an edge $\alpha=(k,l)$. Then
\begin{eqnarray*} 
\mbox{\small $\displaystyle |y_i-y_j|\stackrel{(a)}{\le} \sum_{\alpha\in\gamma^0}|dy_\alpha|
= \sum_{\alpha\in\gamma^0}\sqrt{e_{\alpha}}\frac{|dy_\alpha|}{\sqrt{e_{\alpha}}}
\stackrel{(b)}{\le}
(\sum_{\alpha\in\gamma^0} e_{\alpha} (dy_\alpha)^2)^{\frac12}
(\sum_{\alpha\in\gamma^0} \frac{1}{{e_{\alpha}}})^{\frac12}
\stackrel{(c)}{\le}\sqrt{{\cal E}(y)}\sqrt{D^{\mbox{\small\it sp}}_{ij}}$}
 \end{eqnarray*}
Hence $D^{\mbox{\small\it sp}}_{ij}\ge (y_i-y_j)^2/{\cal E}(y)$ for all $y$, in particular for $y^0$ defined above, showing $D^{\mbox{\small\it sp}}_{ij}\ge D^{\mbox{\small\it com}}_{ij}$. Equality holds iff (a) $y^0$ is monotonously decreasing along the path $\gamma^0$, (b) for all $\alpha\in \gamma^0$, $dy^0_\alpha=c/e_\alpha$ for some constant $c$,  and (c) $dy^0_\alpha e_\alpha=0$ for all $\alpha\notin \gamma^0$. (b), expressing Ohm's law $U=RI$ in the electrical analogy,    holds for $y^0$, and 
(a) and (c) hold for a {\em tree}, that is a graph possessing no closed path.
 
The shortest-path distance is unfocused and irreducible. Seeking  to determine the corresponding  function $g(\lambda)$ involved in Definition  \ref{ndw}, and/or the Schoenberg transformation $\phi(D)$ involved in Definition \ref{edw}, is  however hopeless:

\begin{proposition}
\label{deza}
$D^{\mbox{\small\it sp}}$ is {\em not} a squared Euclidean distance. 
\end{proposition} 

\vspace{-0.2cm}\noindent{\bf Proof:} a  counter-example is provided (Deza and Laurent (1997) p. 83) by the complete bipartite graph $K_{2,3}$ of Figure 1:

\begin{figure} 
\label{k23}
\vspace{-0.5cm}\includegraphics[width=2.4cm]{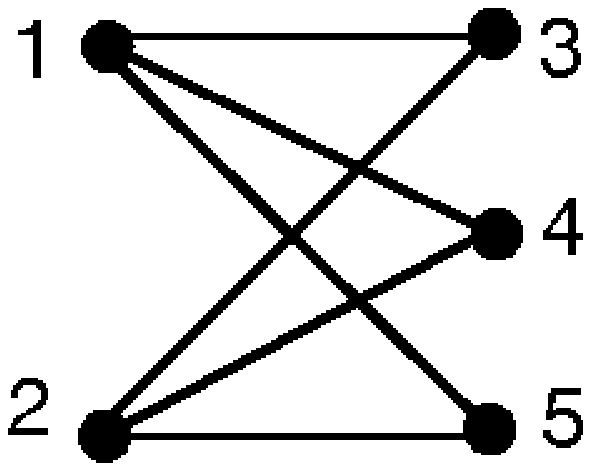}
\hspace{1.3cm}
%\vspace{-1cm}[vspace=-1cm]
\includegraphics[width=3.2cm]{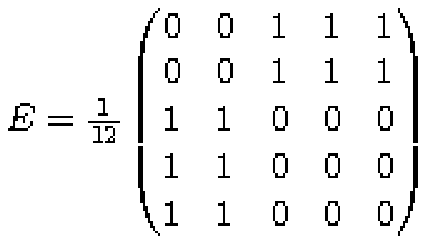}
\hspace{1.2cm}
\includegraphics[width=3.5cm]{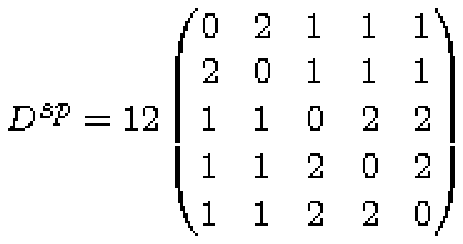}
\caption{Bipartite graph $K_{2,3}$, associated exchange matrix and shortest-path distance
} 
\end{figure} 
The eigenvalues occurring in  Theorem \ref{theomds} are
$\mu_1=3$, $\mu_2=2.32$, $\mu_3=2$, $\mu_4=0$ and  ${\mathbf{\mu_5=-0.49}}$, thus ruling out  the possible squared Euclidean nature of $D^{\mbox{\small\it sp}}$.
 
\subsubsection{The absorption distance}
The choice $g(\lambda)=(1-\rho)/(1-\rho\lambda)$ where $0<\rho<1$ yields the {\em absorption distance}: consider a modified random walk,
where, at each discrete step, a particle at $i$ either undergoes with probability $\rho$ a transition $i\to j$ (with probability $p_{ij}$) or is forever absorbed with probability $1-\rho$ into some additional 
``cemetery" state. The quantities $v_{ij}(\rho)$ 
= ``Òaverage number of visits from $i$ to $j$ before absorption"  obtain as the components of the matrix 
(see e.g. Kemeny and Snell (1976) or Kijima (1997)) 
\begin{displaymath}
V(\rho)=(I-\rho P)^{-1}=(\Pi-\rho E)^{-1}\Pi
\: \: \mbox{with}\: \:  f_iv_{ij}=f_jv_{ji}\: \: \mbox{and}
\: \:  \sum_{i} f_iv_{ij}=\frac{f_j}{1-\rho}
\enspace .
\end{displaymath}
Hence $K=g(\Pi^{-\frac12}E\Pi^{-\frac12})=(1-\rho)\Pi^{\frac12}V\Pi^{-\frac12}$ and 
$B_{ij}=(1-\rho) v_{ij}/f_j$, measuring the ratio of the average number of visits from $i$ to $j$
over its expected value over the initial state $i$. Finally, 
\begin{displaymath}
D_{ij}^{\mbox{\small\it abs}}(\rho)=\frac{v_{ii}(\rho)}{f_i}+\frac{v_{jj}(\rho)}{f_j}-2\frac{v _{ij}(\rho)}{f_j}
\enspace .
\end{displaymath}
By construction, $\lim_{\rho\to 0}D^{\mbox{\small\it abs}}(\rho)=D^{\mbox{\small\it fro}}$ and 
$\lim_{\rho\to 1}(1-\rho)^{-1}D^{\mbox{\small\it abs}}(\rho)=D^{\mbox{\small\it com}}$. Also,
$\lim_{\rho\to 1}D^{\mbox{\small\it abs}}(\rho)\equiv 0$ for a connected graph. 

\subsubsection{The ``sif" distance}
The choice $g(\lambda)=\lambda^2/(1-\lambda)$ is the \textbf{s}implest one insuring an
 {\em  \textbf{i}rreducible and \textbf{f}ocused} squared Euclidean distance. Identity $\lambda^2/(1-\lambda)=1/(1-\lambda)-\lambda-1$ readily yields (wether $D^{\mbox{\small\it dif}}$ is
Euclidean or not)
\begin{displaymath}
D^{\mbox{\small\it sif}}_{ij}=D^{\mbox{\small\it com}}_{ij}-D^{\mbox{\small\it dif}}_{ij}-D^{\mbox{\small\it fro}}_{ij}\enspace .
\end{displaymath}

\section{Numerical experiments}
\label{quatre}
\begin{figure}
\begin{center}
\includegraphics[width=4.4cm]{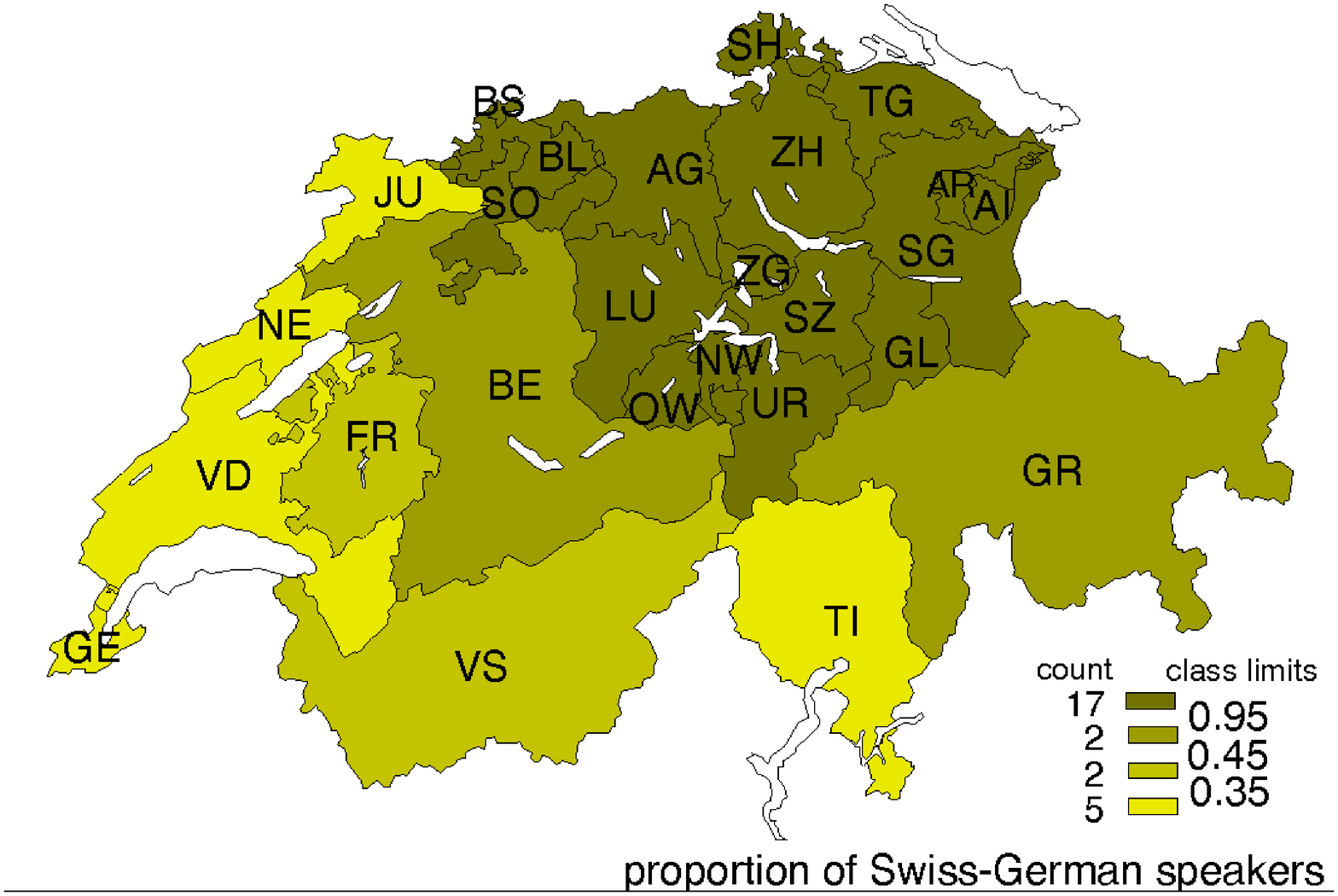}
\includegraphics[width=4.1cm]{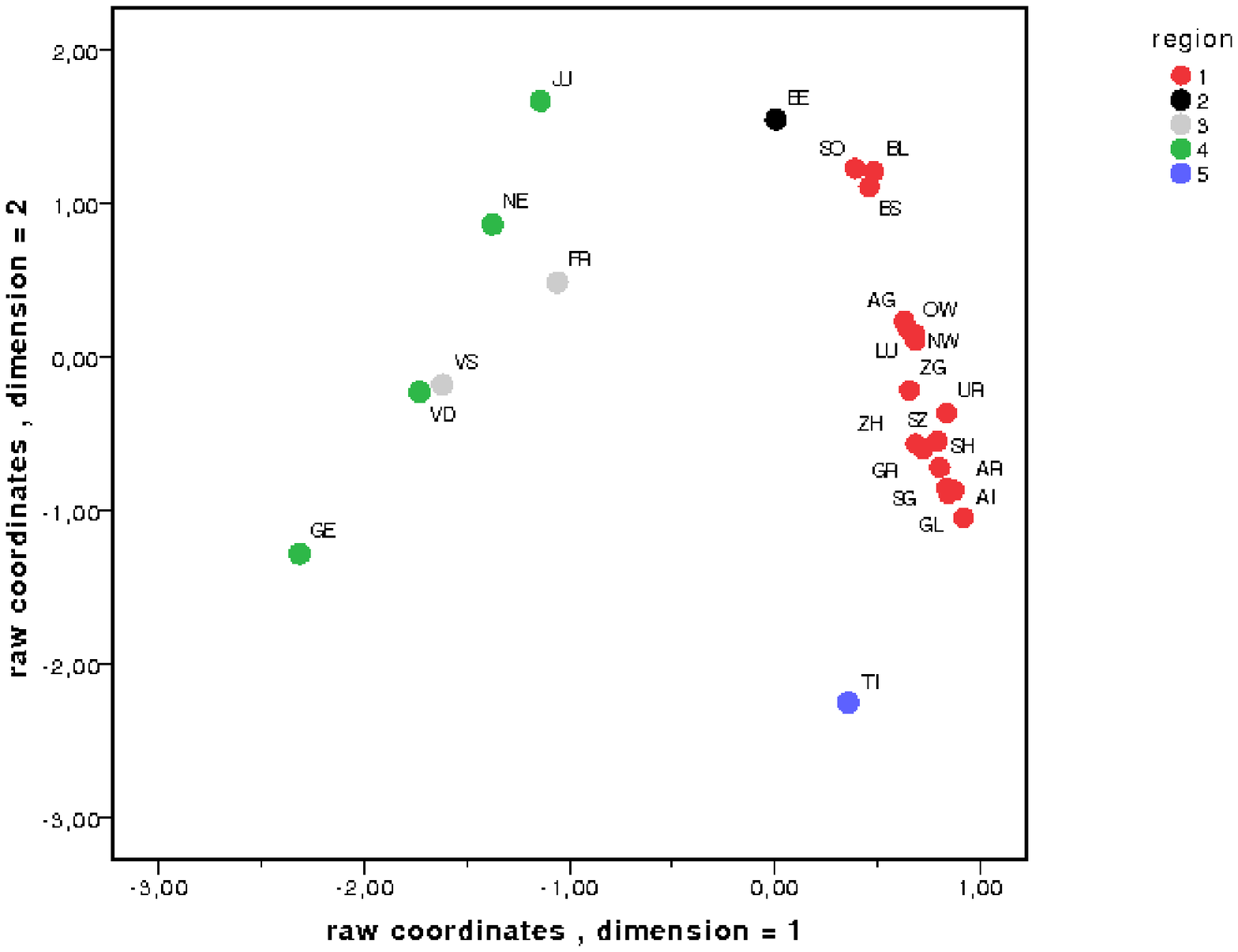}
\includegraphics[width=3.5cm]{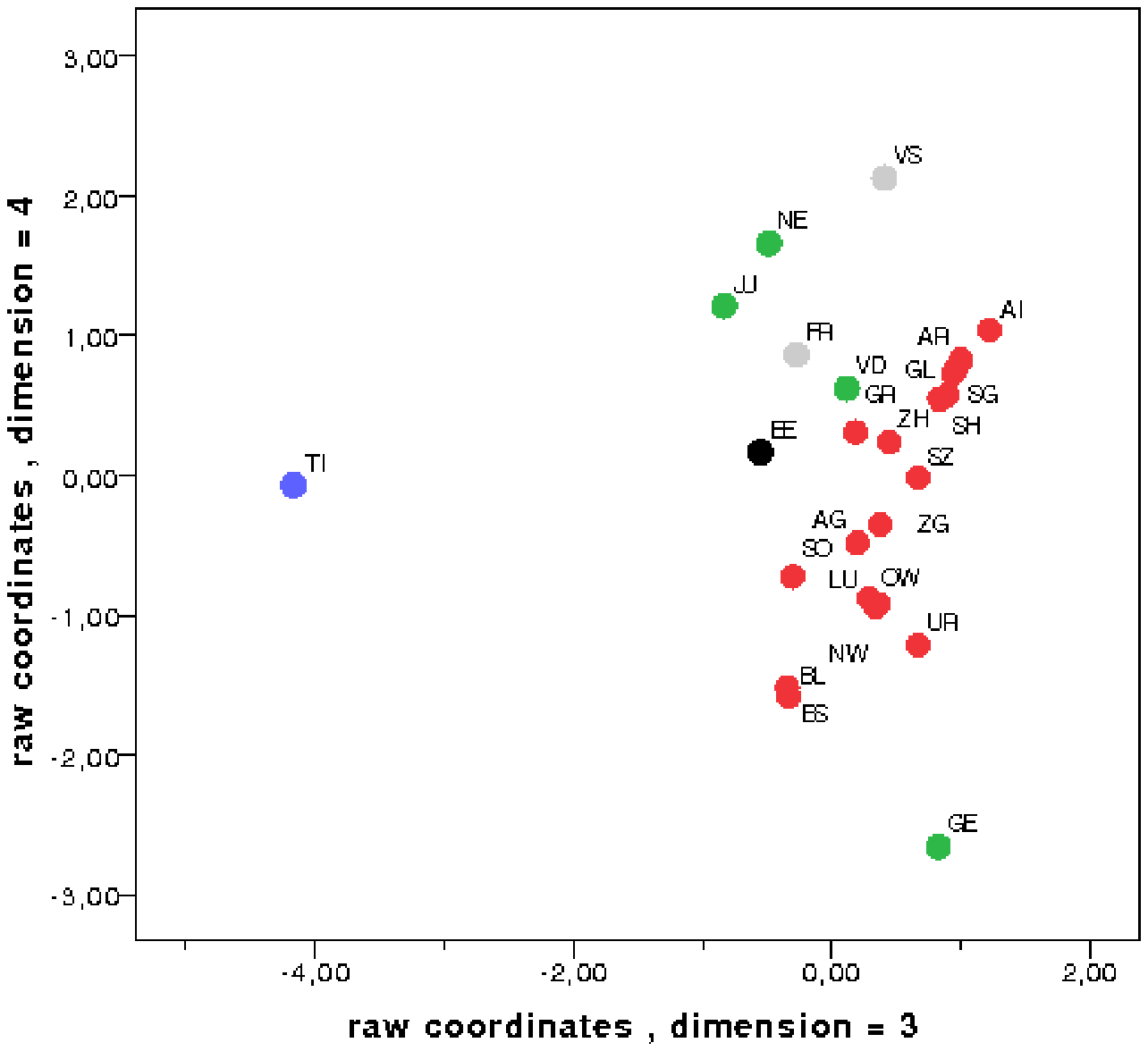}
\caption{Proportion of Swiss-German speakers in the 26 Swiss cantons (left), and raw coordinates $x_{i\alpha}$ associated to the inter-cantonal migrations, in dimensions $\alpha=1,2$ (center) and $\alpha=3,4$  (right). Colours code the linguistic regions, namely: 1 = German, 2 = mainly German, 3 = mainly French, 4 = French and 5 = Italian. The central factorial  map reconstructs fairly precisely the geographical map, and emphasizes the linguistic German-French barrier, known as ``R\"{o}stigraben". The linguistic  isolation of the sole Italian-speaking canton, intensified by the Alpine barrier, is patent.}
\label{fig26}
\end{center}
\end{figure}
 
\subsection{Inter-cantonal migration data}
\begin{figure}
\begin{center}
\includegraphics[width=4cm]{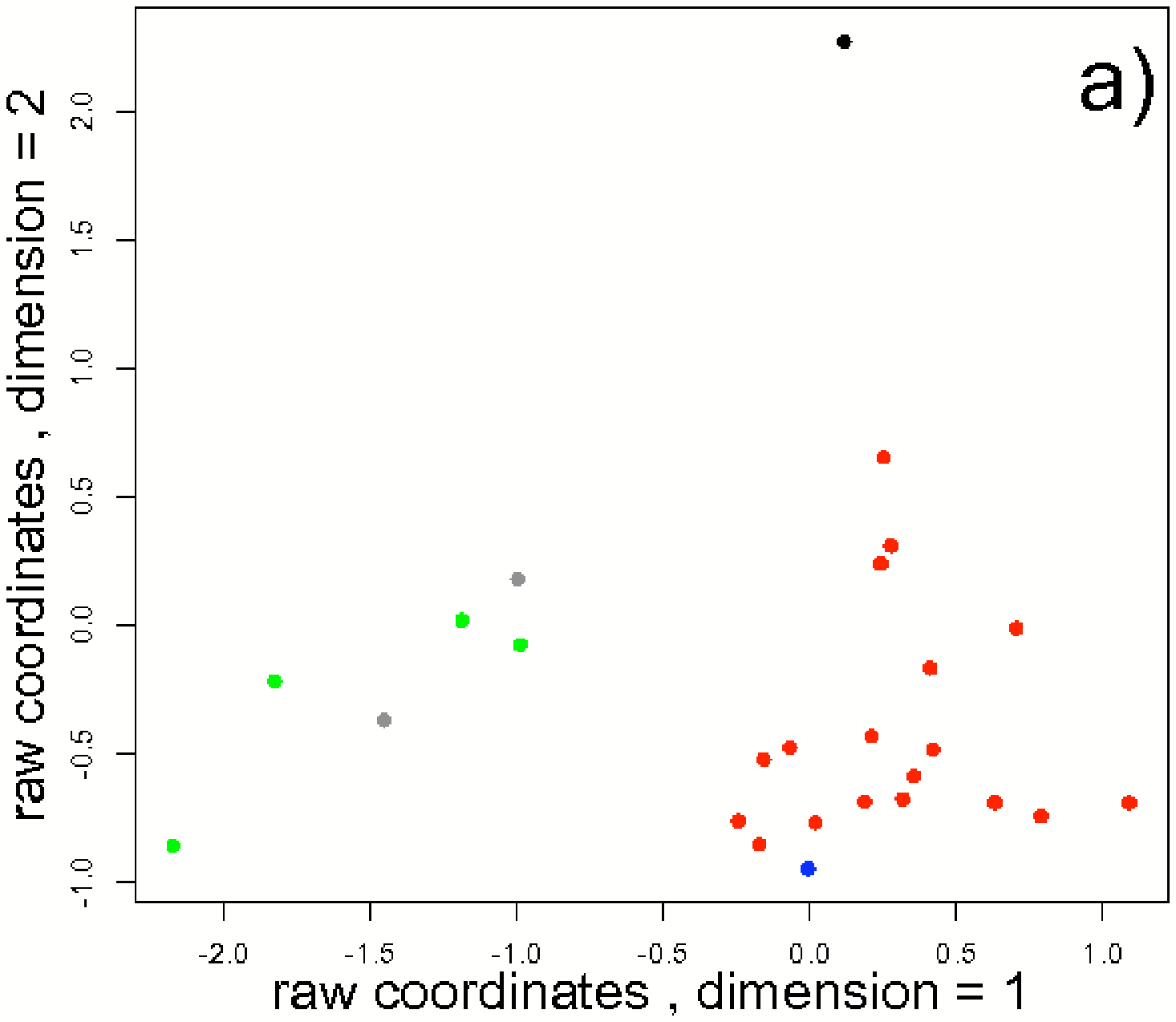}
\includegraphics[width=4cm]{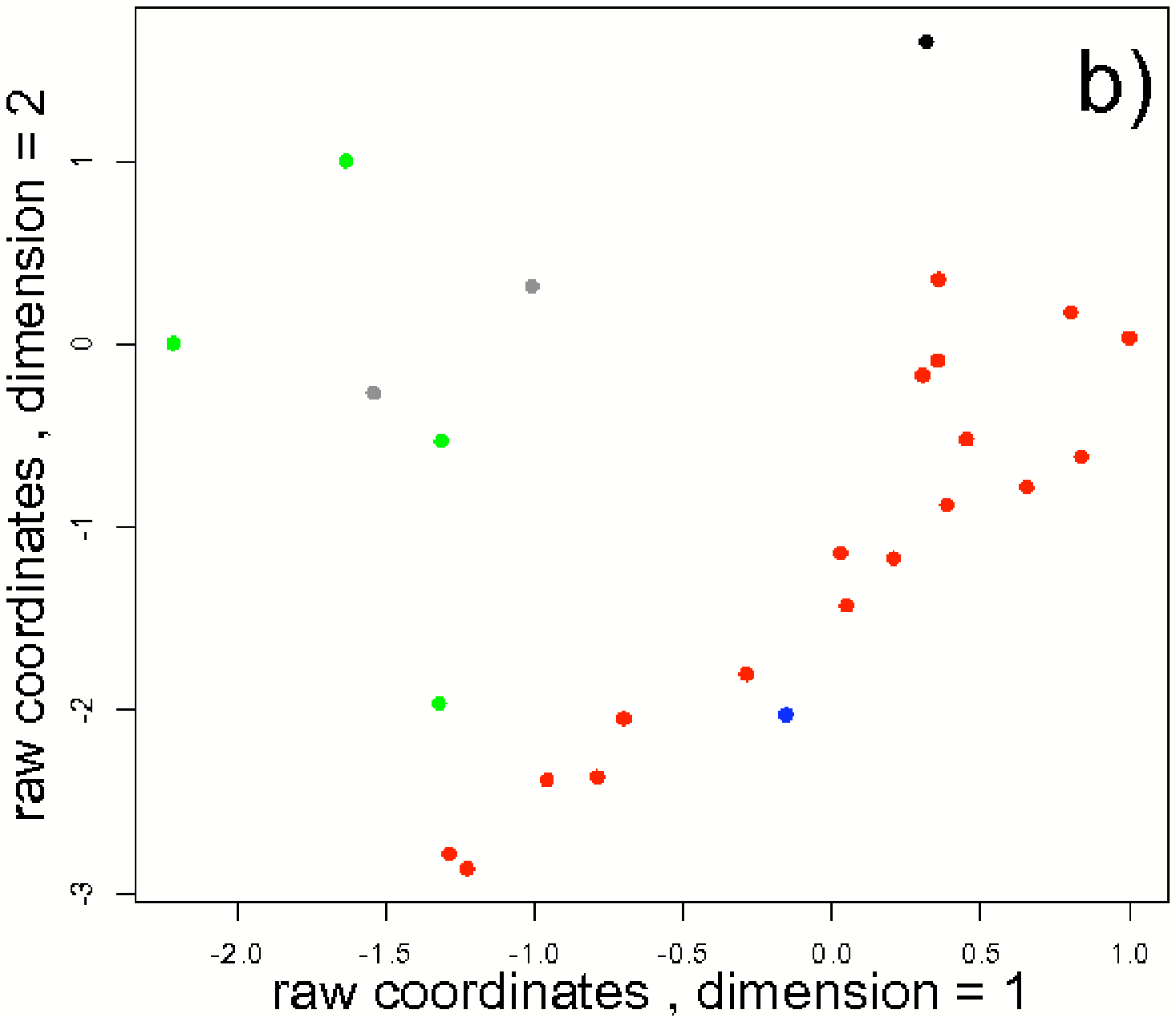}
\includegraphics[width=4cm]{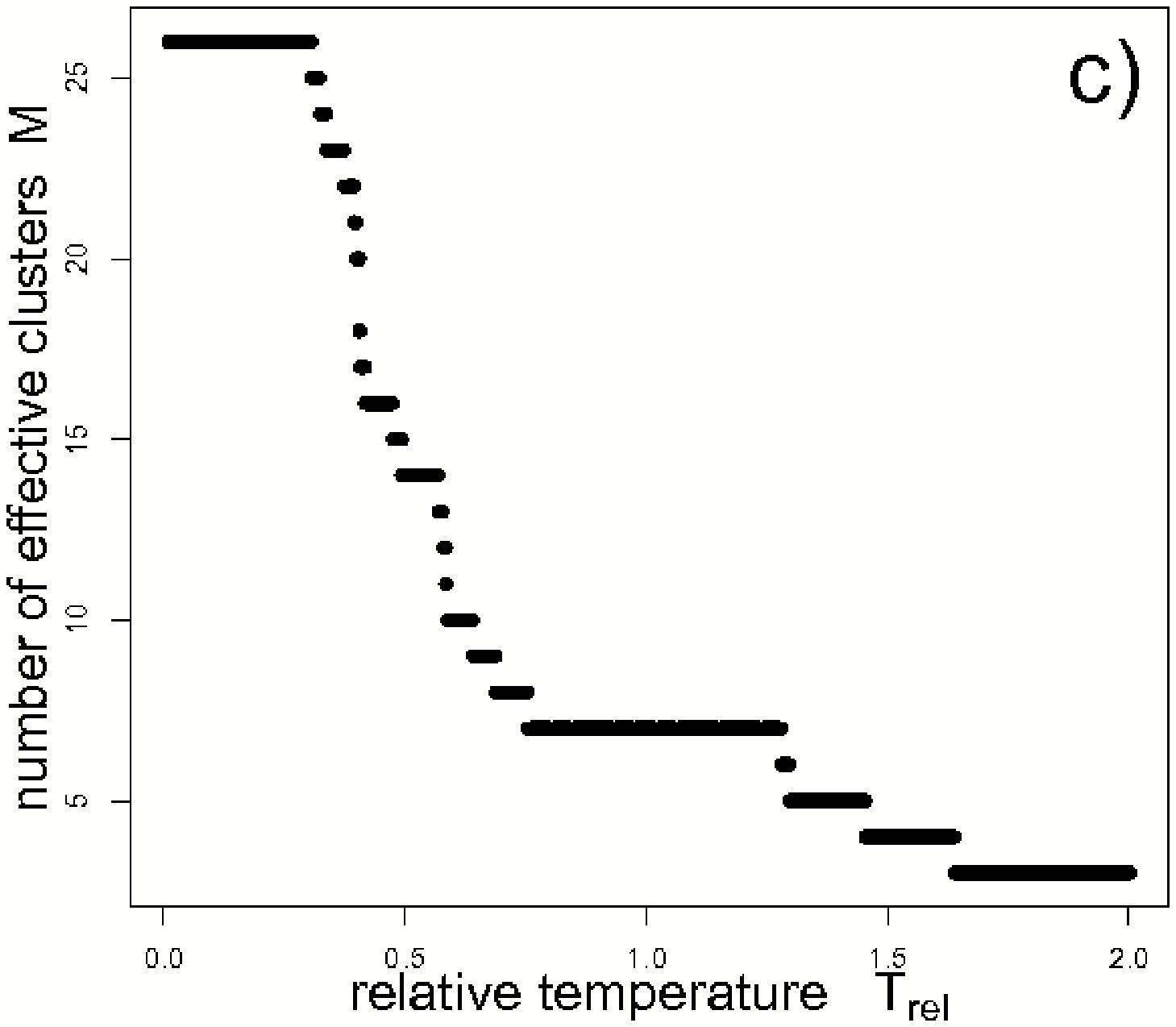}
\includegraphics[width=4cm]{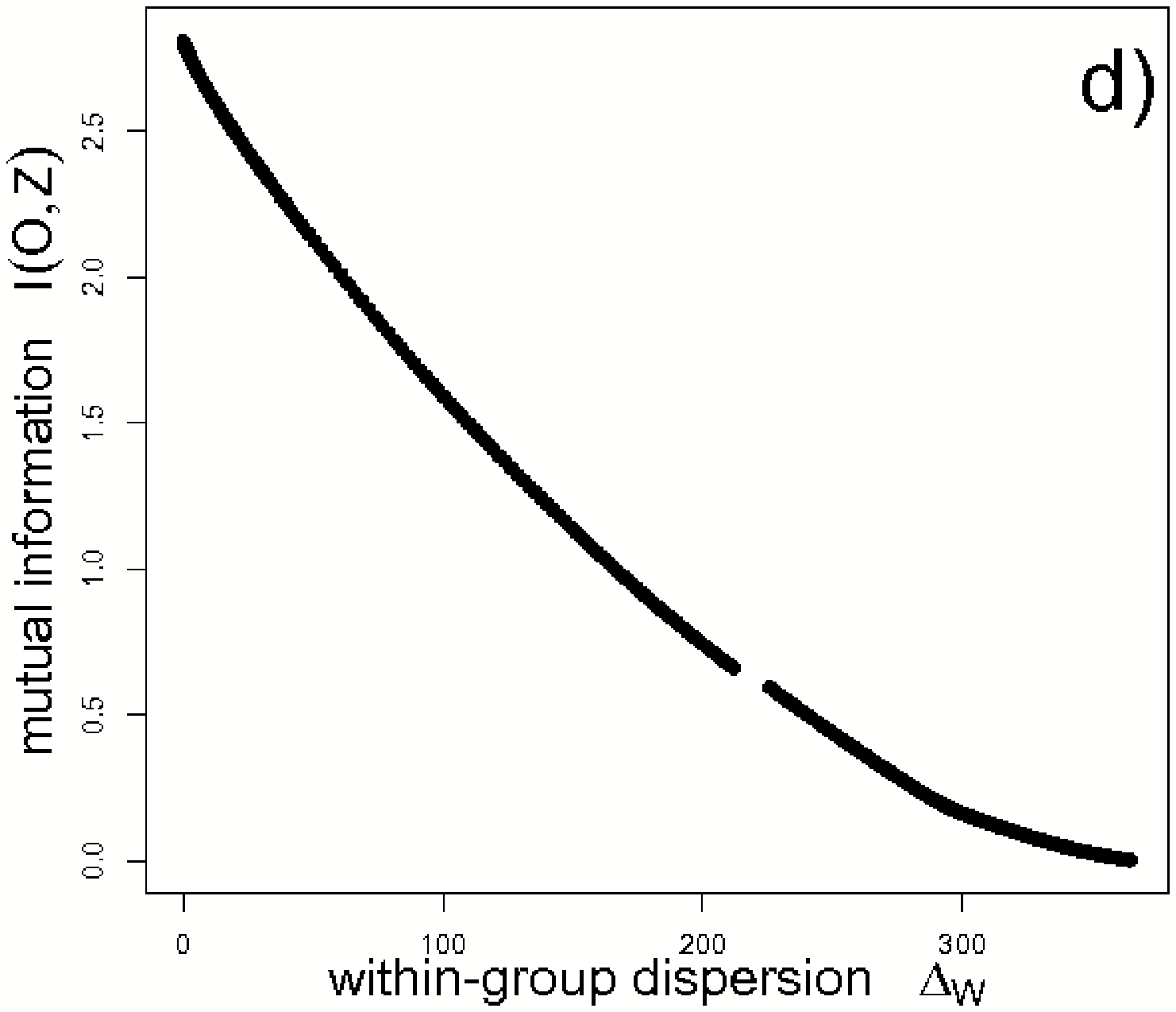}
\includegraphics[width=4cm]{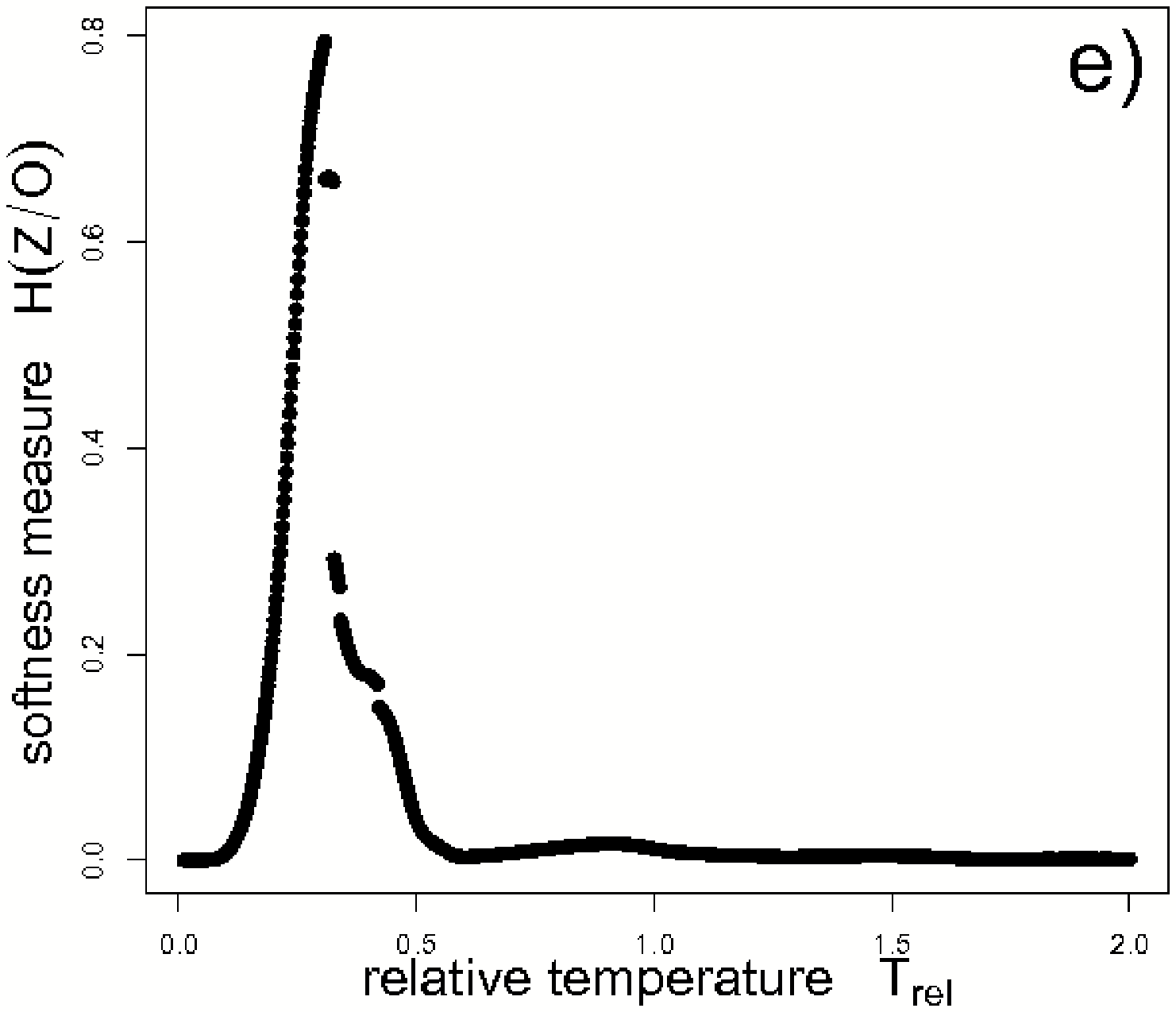}
\includegraphics[width=4cm]{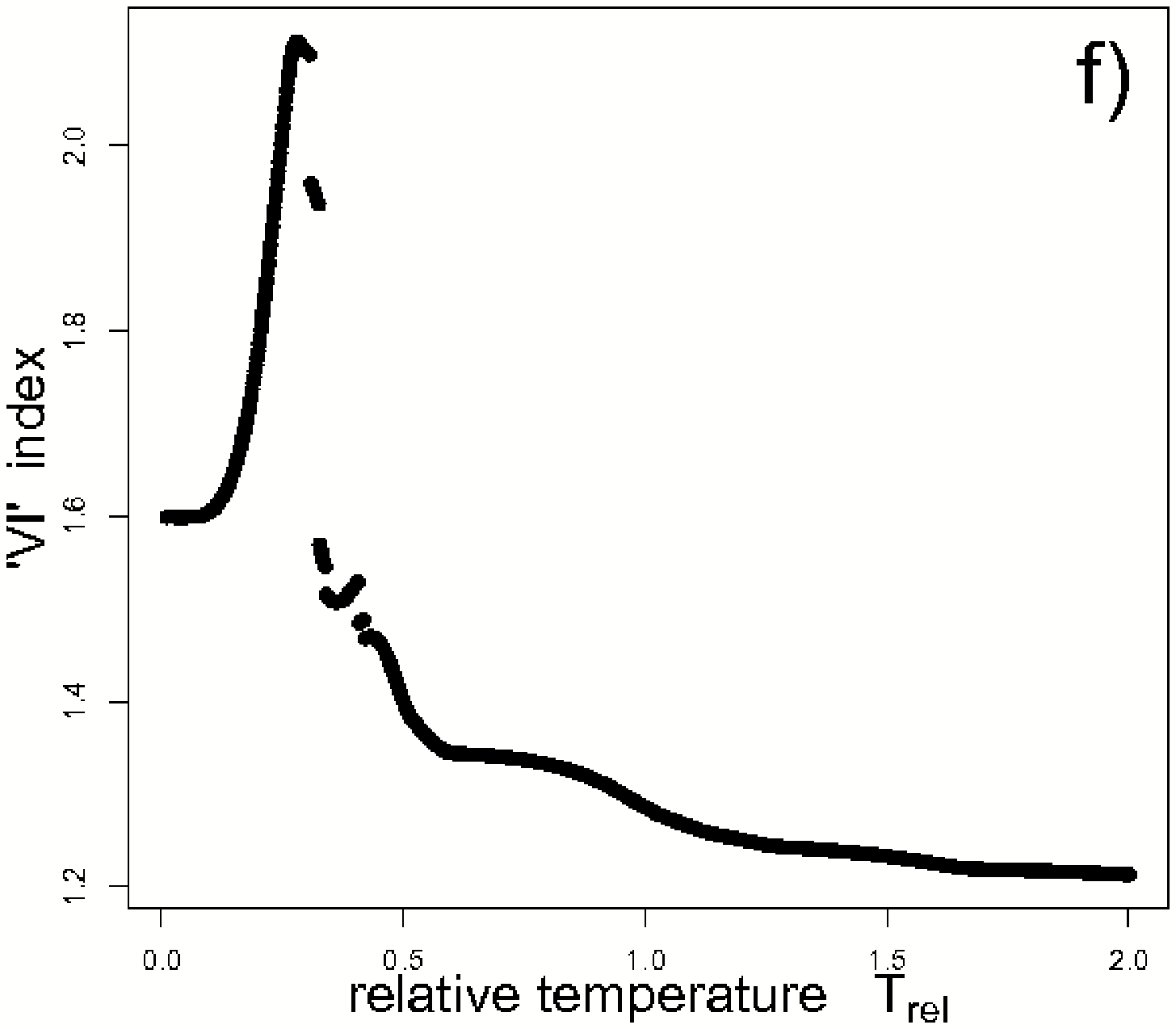}
\caption{Raw coordinates extracted from weighted MDS after applying Schoenberg transformations 
$\tilde{D}=\phi(D^{\mbox{\tiny\it com}})$ with $\phi(D)=D^{0.7}$ (a), and 
$\phi(D)=1-\exp(- b D)$ with $b=1/(4\Delta^{\mbox{\tiny\it com}})$ (b). Decrease of the number of effective groups with the temperature (c); beside the main component, two microscopic groups of size 
$\rho_2=6\cdot 10^{-4}$ and $\rho_3=2\cdot 10^{-45}$ survive at $T_{\mbox{\tiny rel}}=2$. (d) is the so-called {\em rate-distortion function} of Information Theory; its discontinuity at 
$T_{\mbox{\tiny crit}}=0.406$ betrays a {\em phase transition} between a cold regime with numerous clusters and a hot regime with few clusters (Rose et al. 1990; Bavaud 2009). 
Behaviour of the {\em overall softness} $H(Z|O)$ (e) (Section \ref{hsop}) and of the clusters-regions {\em
variation of information} (f) (see text).}
\label{fig6fig}
\end{center}
\end{figure}

The first data set consists of the numbers  $N=(n_{ij})$ of people inhabiting the Swiss canton $i$ in 1980 and the canton $j$ in 1985 ($i,j=1,\ldots, n=26$), with a total count of $6'039'313$ inhabitants, $93\%$ of which are  distributed over the diagonal. $N$ can be made brutally symmetric as $\frac12(n_{ij}+n_{ji})$ or $\sqrt{n_{ij}n_{ji}}$,
or, more gently, by fitting a {\em quasi-symmetric model} (Bavaud 2002), as done here. Normalizing the maximum likelihood estimate yields the exchange matrix $E$. 
Raw coordinates $x_{i\alpha}=u_{i\alpha}/\sqrt{f_i}$ are depicted in  Figure \ref{fig26}. By construction, they do not depend of the form of the function $g(\lambda)$ involved in Definition   \ref{ndw}, but they do depend on the form of the  Schoenberg transformation $\tilde{D}=\phi(D)$ involved in Definition \ref{edw}, where they obtain as solutions of the weighted  MDS algorithm (Theorem \ref{theomds}) on $\tilde{D}$, with unchanged weights $f$ (Figure \ref{fig6fig} (a) and (b)).

Iterating  (\ref{tcl}) from an initial $n\times m$ membership matrix  $Z_{\mbox{\tiny init}}$ (with $m\le n$) at fixed $T$ yields a membership  $Z_0(T)$,  which is by construction a  local minimizer of  the free energy $F[Z,T]$. 
The number $M(Z_0)\le m$ of independent columns of $Z_0$ measures the number of {\em effective groups}: equivalent groups, that is groups whose columns are proportional, could and should be aggregated, thus resulting in $M$ distinct groups, without changing the free energy, since both the intra-group dispersion and the mutual information are aggregation-invariant (Bavaud 2009). In practice, 
groups $g$ and $h$ are  judged as equivalent if their relative overlap (Section \ref{hsop})
 obeys $\theta_{gh}/\sqrt{\theta_{gg}\theta_{hh}}\ge 1-10^{-10}$.

Define  the  relative temperature as $T_{\mbox{\tiny rel}}=T/\Delta$. One expects $M=1$ for $T_{\mbox{\tiny rel}}\gg 1$, and $M=n$ for $T_{\mbox{\tiny rel}}\ll 1$, provided of course that the initial membership matrix  contains at least $n$ columns. We operate a {\em soft hierarchical descendant clustering} scheme, consisting in starting with the identity membership $Z_{\mbox{\tiny init}}=I$ for some $T_{\mbox{\tiny rel}}\ll 1$, iterating  (\ref{tcl}) until convergence, and then aggregating   the equivalent columns in $Z_0(T)$ into  $M$ effective groups. The temperature is then slightly increased, and, choosing the resulting optimum 
$Z_0(T)$ as the new  initial membership, 
(\ref{tcl}) is iterated again, and so forth until the emergence of a  single effective group ($M=1$) in the high  temperature phase $T_{\mbox{\tiny rel}}\ge 1$. 

Numerical experiments (Figure \ref{fig6fig})
actually  conform to the above  expectations, yet with an amazing propensity for tiny groups $\rho_g\ll 1$ to survive at  high temperature, that is  before to be aggregated in the main component. This metastable behaviour is related to the locally optimal nature of the algorithm; presumably unwanted in practical applications, it can be eliminated  by forcing group coalescence if, for instance, $H(Z)$ or $F[Z]-\Delta$ become small enough. 

The softness measure 
of the clustering $H(Z|O)$ is expected to be zero in both temperature limits, since both the identity matrix and the single-group membership matrix are hard. We have attempted to measure the quality of the clustering $Z$ with respect to the regional classification $R$ of Figure \ref{fig26} 
by the ``variation of information" index $H(Z)+H(R)-2I(Z,R)$
proposed by Meila (2005). Further investigations, beyond the scope of this paper, are obviously still  to be conducted in this direction.

The stability of the effective number of clusters around $T_{\mbox{\tiny rel}}=1$ might encourage the choice of the  solution with $M=7$ clusters. Rather disappointingly, the latter turns out (at $T_{\mbox{\tiny rel}}=0.8$, things becoming even worse at higher temperature) to consist of one giant  main component of $\rho_1>0.97$, together with 6 other practically  single-object  groups (UR, OW, NW, GL, AI, JU), totalizing less than three percent of the total mass (see also Section \ref{conconc}).

\subsection{Commuters data}
The second data set counts the number of commuters $N=n_{ij}$ between the $n=892$  French speaking Swiss communes, living in commune $i$ and working in commune $j$ in 2000. A total of $733'037$ people are involved, $49\%$ of which are distributed over the diagonal. As before, the 
exchange matrix $E$ is obtained after fitting a quasi-symmetric model to $N$.  
\begin{figure}[h]
\begin{center}
\includegraphics[width=4cm]{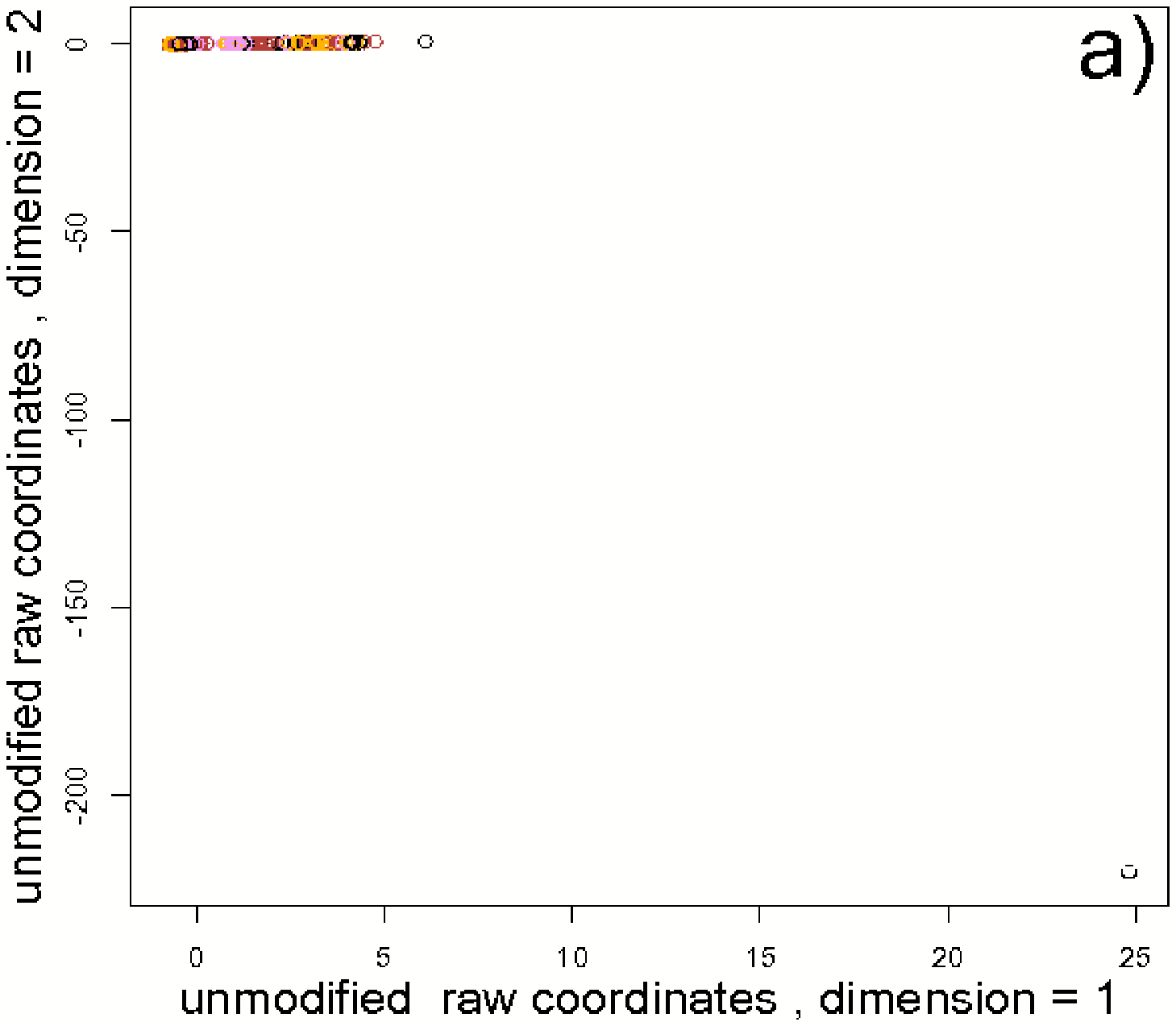}
\includegraphics[width=4cm]{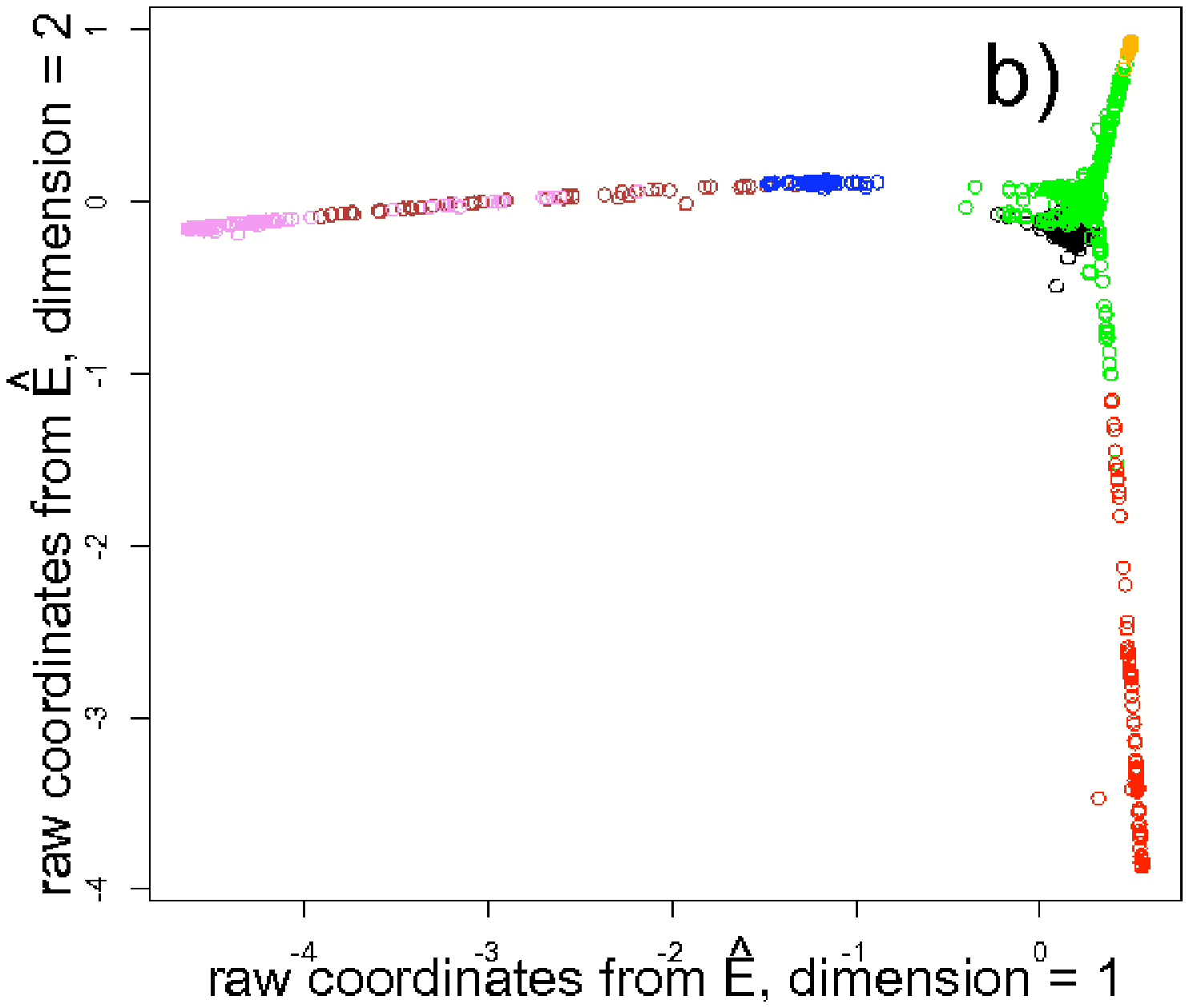}
\includegraphics[width=4cm]{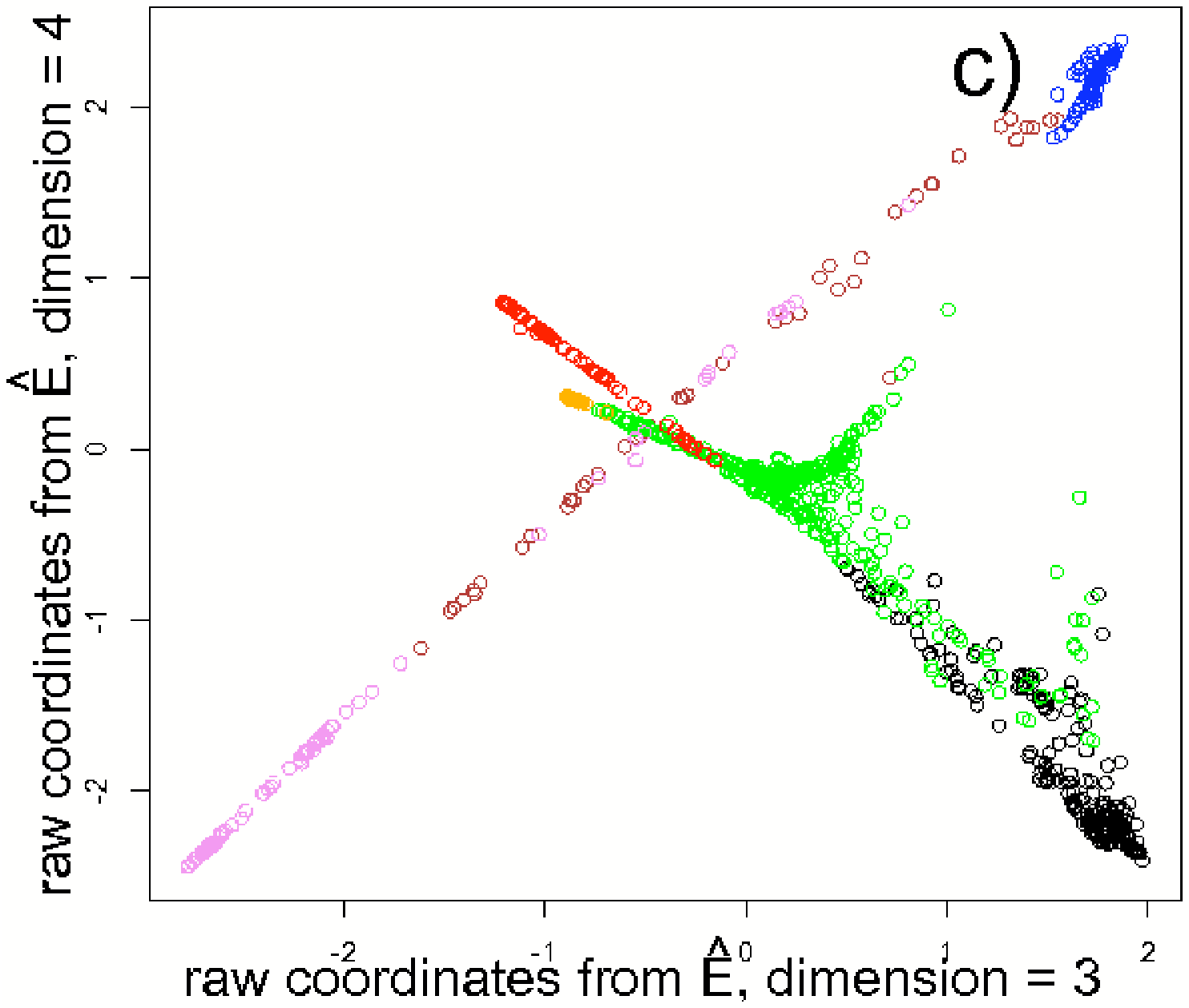}
\includegraphics[width=4cm]{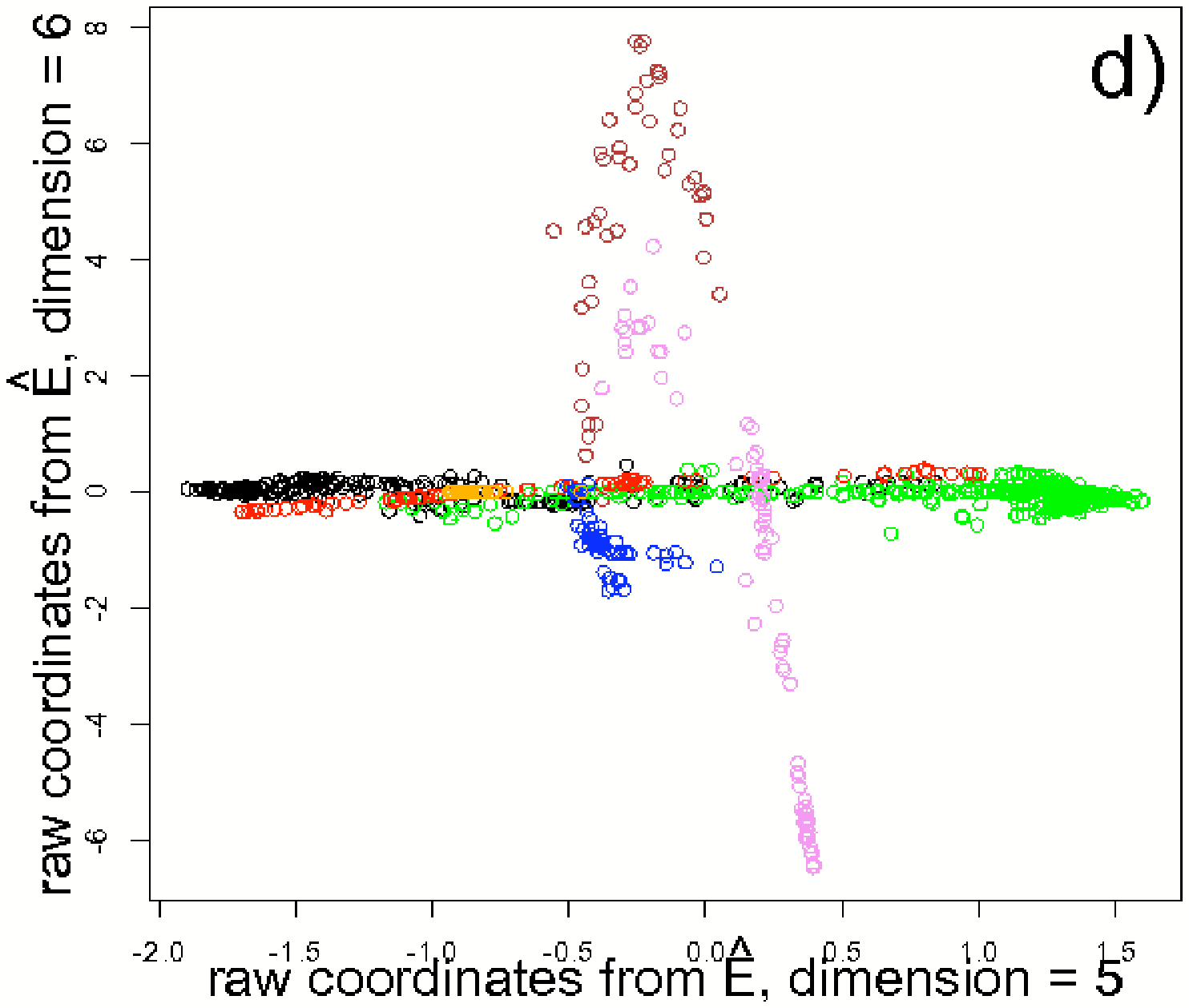}
\includegraphics[width=4cm]{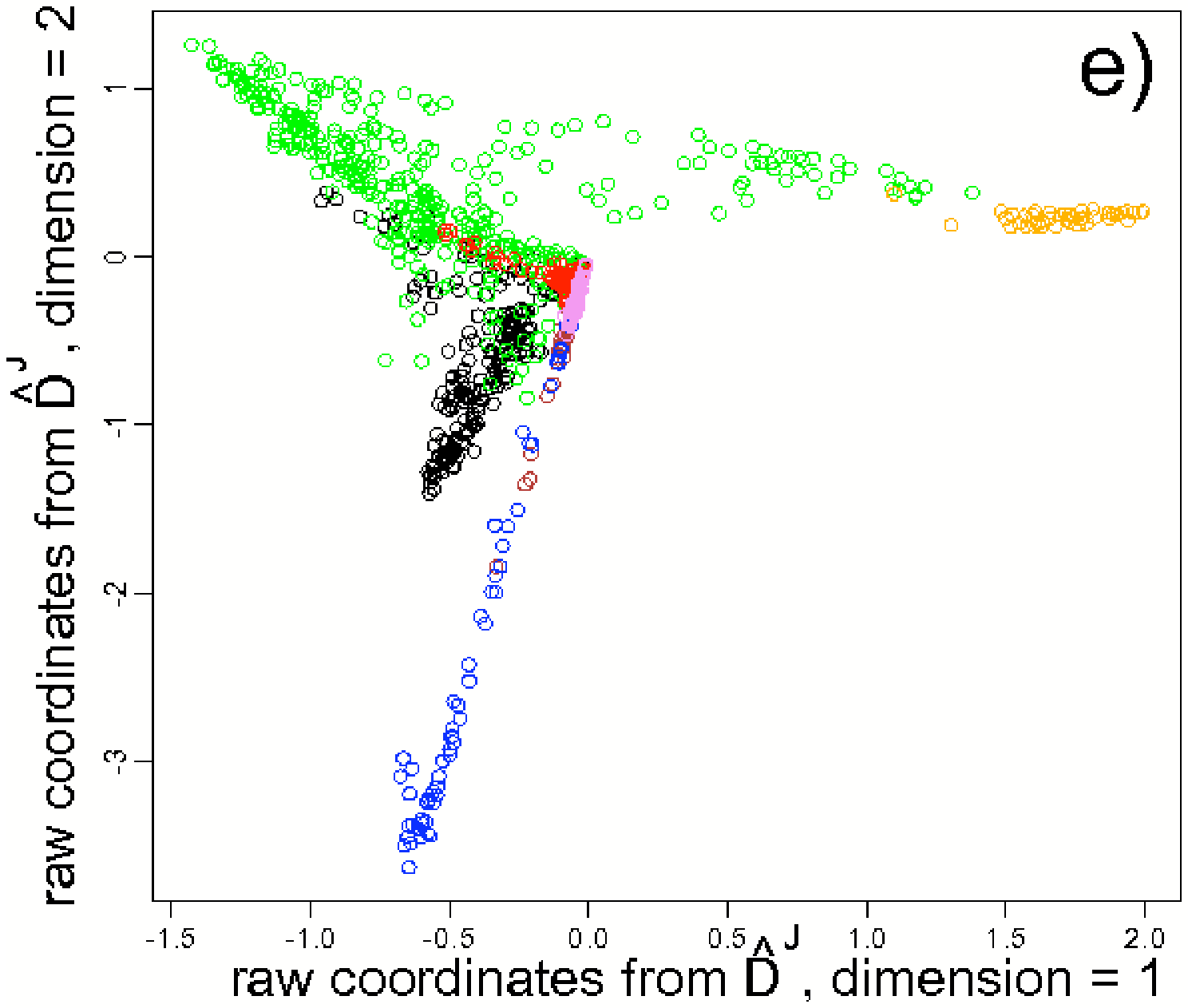}
\includegraphics[width=4cm]{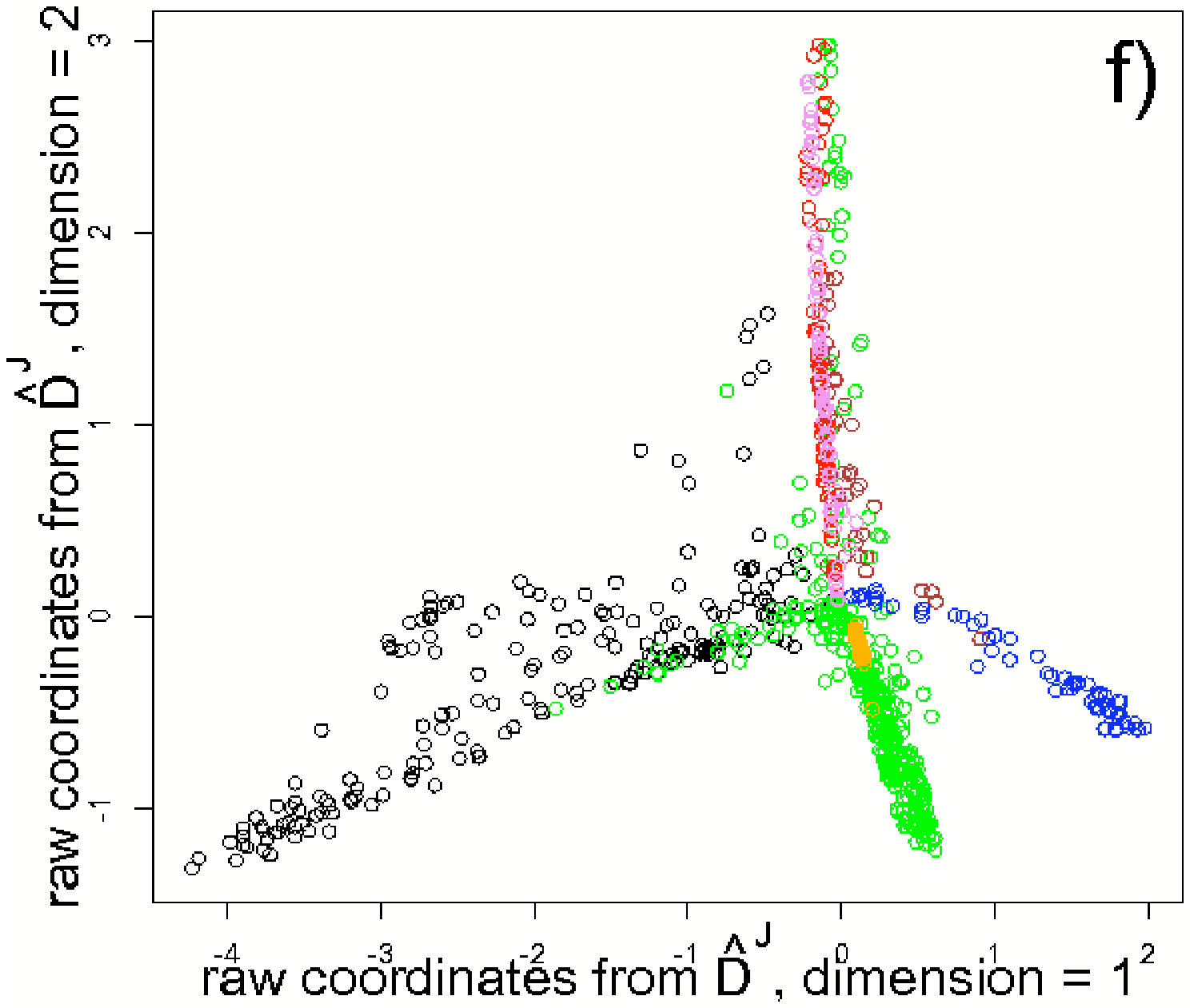}
\caption{Raw coordinates associated to the unmodified exchange matrix $E$ are unable to approximate the geographical  map (a), in contrast to (b),  (c) and (d), based upon the diagonal-free exchange matrix $\hat{E}$. Colours code the cantons, namely BE=brown,  FR=black, GE=orange, JU=violet,  NE=blue, VD=green, VS=red. In particular, the  central position of VD (compare with Figure \ref{fig26}) is confirmed. (e) and (f) represent the low-dimensional  coordinates obtained by MDS from $\hat{D}^{\mbox{\tiny \it jump}}$ (\ref{epsi}).}
\label{fig892}
\end{center}
\end{figure}
  The first two dimensions $\alpha=1,2$ of the raw coordinates  $x_{i\alpha}=u_{i\alpha}/\sqrt{f_i}$  are depicted in Figure  \ref{fig892} a). The objects cloud consists of all the communes (up, left) except a single one (down, right), namely ``Roche d'Or" (JU), containing 15 active inhabitants, 13 of which work in Roche d'Or. Both the very high value of the  proportion of stayers $e_{ii}/f_i$ and the low value of the weight $f_i$ make Roche d'Or (together with other communes, to a lesser extent) quasi-disconnected from the rest of the system, hence  producing, in accordance to the theory,  eigenvalues as high as $\lambda_1=.989$, 
 $\lambda_2=.986$,   ... , $\lambda_{30}>.900$...
 
 Theoretically flawless as is might be, this behavior stands as a complete geographical failure. As a matter of fact, commuters (and migration)-based  graphs are {\em young},  that is $E$ is much closer to its short-time limit $E^{(0)}$ than to its equilibrium value $E^{(\infty)}$. 
 %They are in addition  strongly {\em irregular},  that is their weights, far from uniform, follow a Zipf 
 % power-law with a large exponent, with an  ``irregularity index"  (see Section \ref{hsop}) value of
 %   $\ln n- H(O)=1.84$, and $0.45$ for the migration data. 
Consequently,  diagonal components are huge and equivalent vertices in the sense of Definition \ref{fod} cannot exist: for $k=i\neq j$,  
the proportion of stayers $e_{ii}/f_i$ is large,   while $e_{ij}/f_j$ is not.  

Attempting  to consider the {\em Laplacian} $E-E^{(0)}$ instead of $E$ does not improve the situation:  both matrices indeed generate the same eigenstructure, keeping the order of eigenvalues unchanged. A brutal, albeit more effective strategy consists in plainly destroying the diagonal exchanges, that is by replacing $E$ by the {\em diagonal-free exchange  matrix} $\hat{E}$,   with components  and 
associated weights  
\begin{displaymath}
\hat{e}_{ij}=\frac{e_{ij}-\delta_{ij}e_{ii}}{1-\sum_k e_{kk}}
\qquad\qquad\qquad \hat{f}_i=\frac{f_i-e_{ii}}{1-\sum_k e_{kk}}\enspace .
\end{displaymath}
Defining $\hat{E}$  as the new exchange matrix  yields (Sections \ref{deux} and \ref{trois}) new weights $\hat{f}$, eigenvectors $\hat{U}$, eigenvalues $\hat{\Lambda}$ (with $\hat{\lambda}_n=0$), raw coordinates $\hat{X}$ and distances $\hat{D}$, as illustrated in Figure \ref{fig892} b), c) and d). 

However, an example of equivalent nodes in the  sense of   Definition \ref{fod} is still unlikely to be found, since $0=\hat{e}_{ii}/\hat{f}_i\neq \hat{e}_{ij}/\hat{f}_j$ in general.
A weaker concept of equivalence consists in comparing $i\neq j$ by means of their transition probabilities towards the {\em other} vertices $k\neq i,j$, that is by means of the   Markov chain conditioned  to the event that the next state is {\em different}. Such Markov 
transitions approximate the so-called {\em jump}  process, if existing (see e.g. Kijima (1997) or  Bavaud (2008)).  Their  associated exchange matrix is precisely given by $\hat{E}$.  

 \begin{definition}[Weakly equivalent vertices; weakly focused distances]
\label{wod}
Two distinct vertices $i$ and $j$ are {\rm weakly equivalent}, noted $i\stackrel{w}{\sim} j$, if $\hat{e}_{ik}/\hat{f}_i= \hat{e}_{jk}/\hat{f}_j$ for all $k\neq i,j,$.
A distance is {\rm weakly focused}  if $D_{ij}=0$ whenever $i\stackrel{w}{\sim} j$. 
\end{definition}

By construction, the following ``jump" distance is squared Euclidean and weakly focused: 
\begin{equation}
\label{epsi}
\hat{D}^{\mbox{\small\it jump}}_{ij}=\sum_{k\: |\: k\neq i,j}\hat{f}_k(\frac{\hat{e}_{ik}}{\hat{f}_i\hat{f}_k}-
\frac{\hat{e}_{jk}}{\hat{f}_j\hat{f}_k})^2=\sum_k\frac{1}{\hat{f}_k}(\frac{\hat{e}_{ik}}{\hat{f}_i}-
\frac{\hat{e}_{jk}}{\hat{f}_j})^2-\frac{\hat{e}_{ij}^2}{\hat{f}_i\hat{f}_j}(\frac{1}{\hat{f}_i}+\frac{1}{\hat{f}_j})
\enspace .
\end{equation}
The restriction $k\neq i,j$ in (\ref{epsi}) complicates  the expression of 
$D^{\mbox{\small\it jump}}$ in terms of the eigenstructure $(\hat{U},\hat{\Lambda})$, and the existence of 
raw coordinates $\hat{x}_{i\alpha}$, adapted to the diagonal-free case, and justified by an analog of Proposition \ref{equ}, remains open.  In any case, jump distances (\ref{epsi}) are well defined, and yield low-dimensional coordinates of the 892 communes by weighted MDS (Theorem \ref{theomds}) with weights $\hat{f}$, as illustrated in Figure \ref{fig892} e) and f).

  \begin{figure}
\begin{center}
\includegraphics[width=4cm]{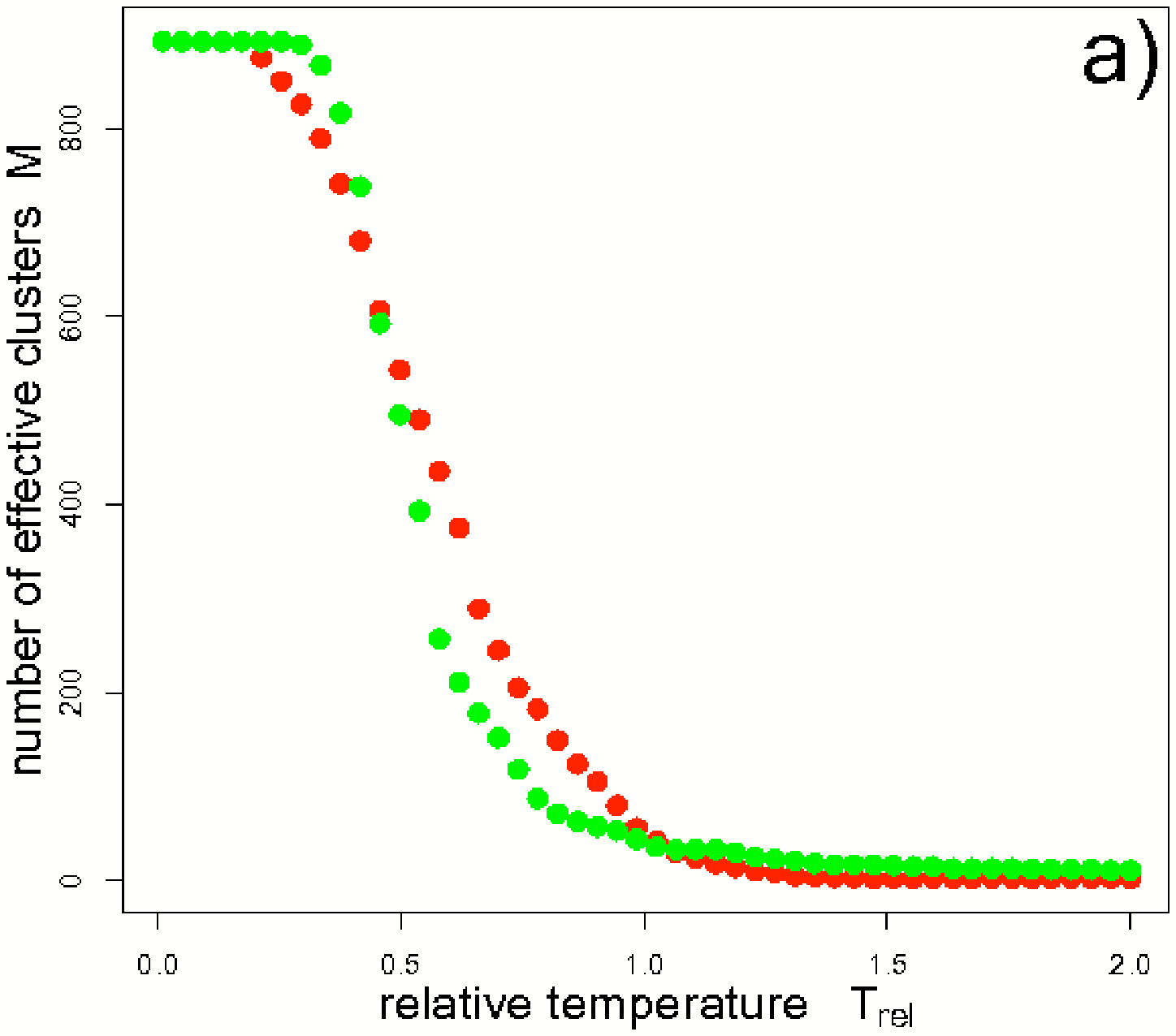}
\includegraphics[width=4cm]{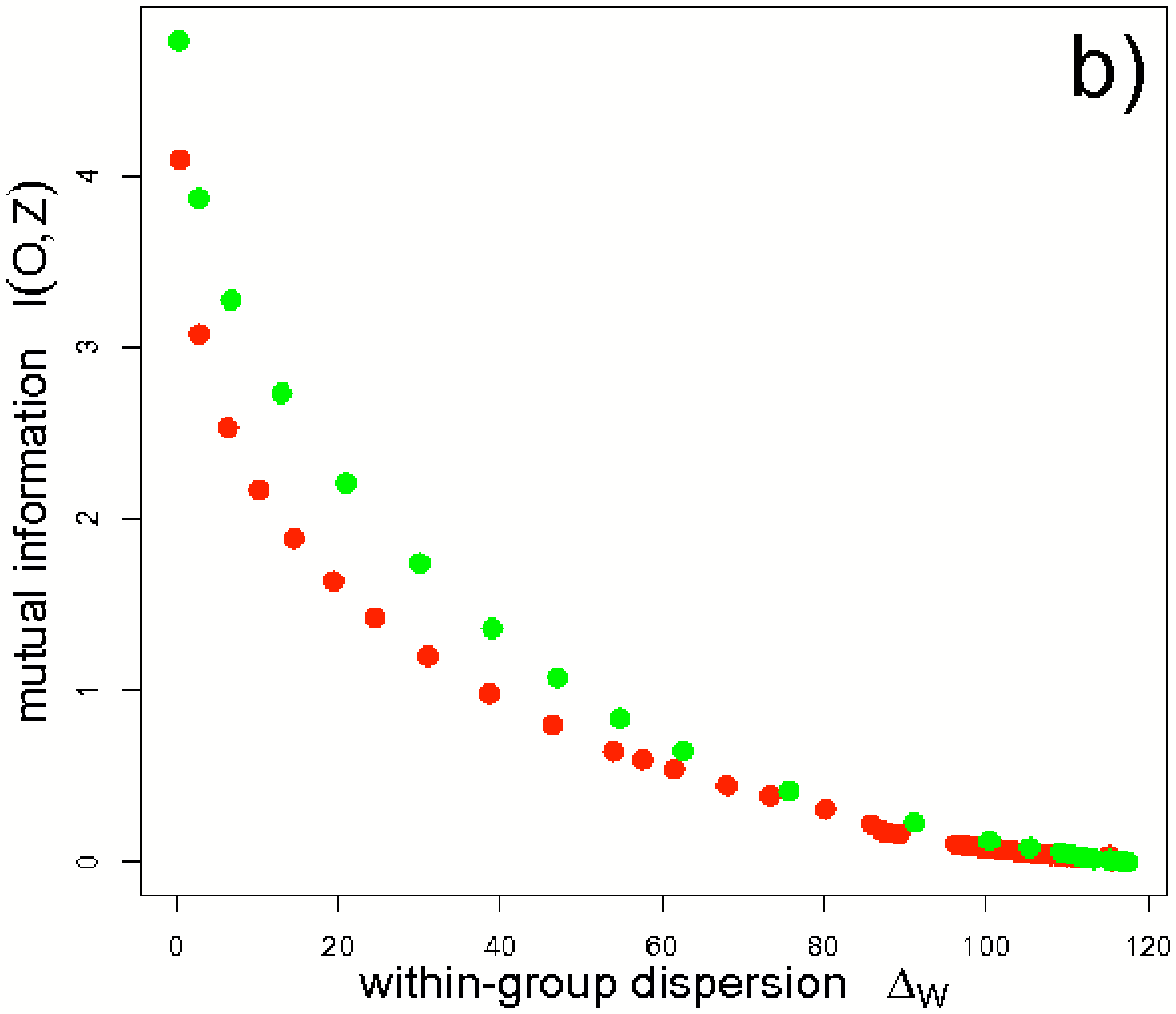}
\includegraphics[width=4cm]{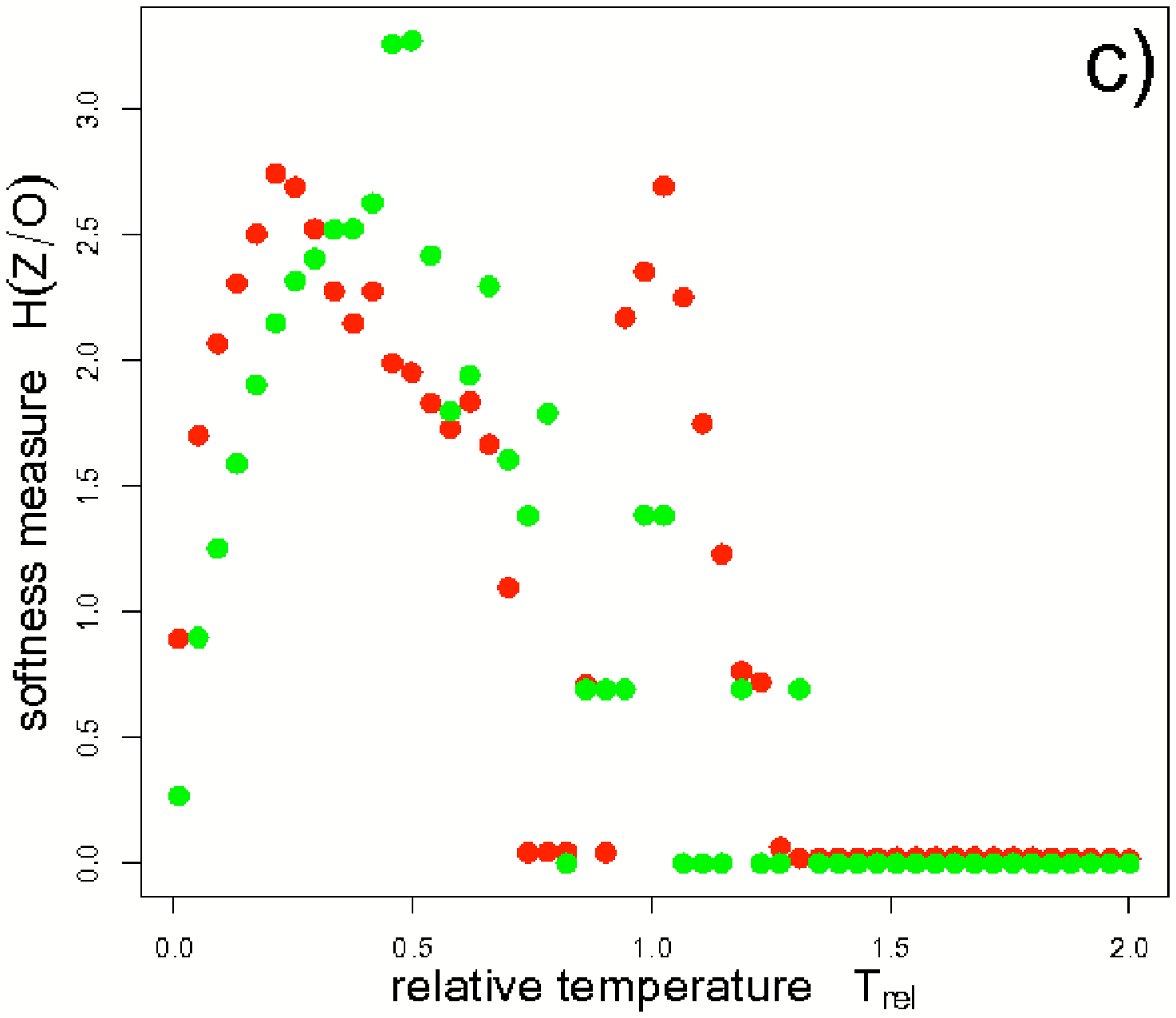}
\caption{Comparison between the clustering obtained from $\hat{D}^{\mbox{\tiny \it sif}}$ (in red) and
$\hat{D}^{\mbox{\tiny \it jump}}$ (in green): evolution of the number of effective clusters with the temperature (a), rate-distortion function (b) and overall softness measure (c). In (b), 
$\hat{\Delta}^{\mbox{\tiny \it jump}}$ has been multiplied by a factor  five to fit to the scale.}
\label{fig892B}
\end{center}
\end{figure}

\section{Conclusion}
\label{conconc}
%The first results confirm the theoretical coherence and the numerical feasibility of the whole procedure. %Yet, further investigations are    certainly  required: in particular, the precise role   that  the diagonal 
Our first numerical results confirm the theoretical coherence and the tractability of the clustering procedure presented in this paper. Yet, further investigations are    certainly  required: in particular, the precise role   that  the diagonal 
components of the exchange matrix should play into the construction of  distances on graphs deserves to be thoroughly elucidated. Also, the presence of fairly small clusters in the clustering solutions of Section 
\ref{quatre}, from which the normalized cut algorithm {\tt Ncut} was supposed to prevent, should be fully understood. Our present guess is that small clusters are inherent to the spatial nature of the data under consideration: elongated and connected clouds as those of Figure \ref{fig892} {\em cannot} miraculously  split  into well-distinct groups, irrespectively of the details of the clustering algorithm (classical chaining problem). This being said,  squared Euclidean are closed under addition and convex mixtures. Hence, an elementary yet principled remedy could simply consist in adding {\em spatial squared Euclidean distances} to the {\em flow-induced distances} investigated in the present contribution.

\end{document}